\documentclass{article}
\usepackage{iftex}
\ifPDFTeX
  \PackageError{main}{XeLaTeX or LuaLaTeX is required}{Select XeLaTeX in your editor or Overleaf project settings so UTF-8 Chinese text can be compiled.}
\fi
\usepackage[UTF8,scheme=plain,fontset=fandol]{ctex}
\usepackage{iclr2026_conference}
\usepackage{amsmath,amsfonts,bm}

\def\eqref#1{equation~\ref{#1}}

\def\1{\bm{1}}

\DeclareMathAlphabet{\mathsfit}{\encodingdefault}{\sfdefault}{m}{sl}
\SetMathAlphabet{\mathsfit}{bold}{\encodingdefault}{\sfdefault}{bx}{n}

\usepackage{hyperref}
\definecolor{linkblue}{RGB}{85,137,201}
\hypersetup{
  colorlinks=true,
  linkcolor=linkblue,
  citecolor=linkblue,
  urlcolor=linkblue
}
\usepackage{graphicx}
\usepackage{url}
\usepackage{booktabs}
\usepackage{colortbl}
\usepackage{tabularx}
\usepackage{array}
\usepackage{float}
\usepackage{flafter}
\usepackage{placeins}
\usepackage{wrapfig}
\usepackage{bbding}
\usepackage{etoolbox}

\definecolor{totalmetricgray}{gray}{0.93}
\definecolor{projectlinkpink}{RGB}{229,43,121}

\AtBeginEnvironment{figure}{\setlength{\abovecaptionskip}{4pt}}
\AtBeginEnvironment{figure*}{\setlength{\abovecaptionskip}{4pt}}

\makeatletter
\newcommand{\compacttablecaption}[1]{%
    \begingroup
    \renewcommand{\@makecaption}[2]{%
        \vskip\abovecaptionskip
        \hb@xt@\hsize{\hfil
            {\fontsize{7.3pt}{7.2pt}\selectfont ##1: ##2}%
        \hfil}%
        \vskip\belowcaptionskip
    }%
    \caption{#1}%
    \endgroup
}
\newcommand{\ablationtablecaption}[1]{%
    \begingroup
    \renewcommand{\@makecaption}[2]{%
        \vskip\abovecaptionskip
        \hb@xt@\hsize{\hfil
            {\fontsize{8.5pt}{9.5pt}\selectfont ##1: ##2}%
        \hfil}%
        \vskip\belowcaptionskip
    }%
    \caption{#1}%
    \endgroup
}
\makeatother

\newcommand{\methodname}{\textsc{PhiZero}}
\newcommand{\Methodname}{\methodname}
\newcommand{\titlelogo}{%
    \raisebox{-0.23\height}{%
        \includegraphics[
            height=1.25em
        ]{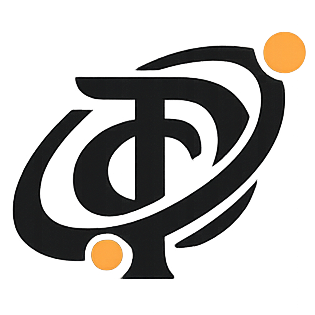}%
    }%
}
\title{{
\titlelogo\hspace{0.25em}\methodname{}: A World Model Built Around Physical Language}}

\newcommand{\projectleadermark}{\textsuperscript{\normalfont\textdagger}}
\newcommand{\correspondingmark}{\textsuperscript{\normalfont\Envelope}}
\newcommand{\independentmark}{\textsuperscript{\normalfont\S}}
\newcommand{\physicalscopefootnote}{%
    \begingroup
    \renewcommand{\thefootnote}{\fnsymbol{footnote}}%
    \footnote{Here, \emph{physical} refers broadly to physical world, rather than to a symbolic system of physical laws.}%
    \addtocounter{footnote}{-1}%
    \endgroup
}

\author{%
\raisebox{-20pt}{%
\makebox[\dimexpr\textwidth-2\tabcolsep\relax][c]{%
\begin{tabular}{c}
Shuyao Shang
\quad Yuqi Wang\projectleadermark\independentmark
\quad Ruopeng Gao\independentmark
\quad Xu Chen\independentmark
\\
Tieniu Tan
\quad Lue Fan\correspondingmark
\quad Zhaoxiang Zhang\correspondingmark
\\[0.5em]
{\normalfont\small NLPR, Institute of Automation, Chinese Academy of Sciences (CASIA)}
\\[0.65em]
{\normalfont\normalsize
\hspace{0.2em}%
{\hypersetup{pdfborder={0 0 0}}%
 \href{https://Phi-Zero.github.io/}{%
     \textcolor{projectlinkpink}{\textit{https://Phi-Zero.github.io/}}%
 }}}
\end{tabular}%
}%
}%
}

\iclrfinalcopy
\begin{document}

\maketitle
\begin{NoHyper}
\begingroup
\renewcommand{\thefootnote}{\textdagger}
\footnotetext{Project Leader.\quad
\textsuperscript{\Envelope}Corresponding Authors.\quad
\textsuperscript{\S}Independent Researcher.}
\endgroup
\end{NoHyper}
\setcounter{footnote}{0}
\pagestyle{fancy}
\fancyhead{}
\fancyhead[L]{Preprint}
\fancypagestyle{plain}{%
    \fancyhf{}%
    \fancyhead[L]{Preprint}%
    \fancyfoot[C]{\thepage}%
}
\thispagestyle{fancy}

\vspace{-2em}
\begin{figure}[H]
    \centering
    \includegraphics[width=\linewidth]{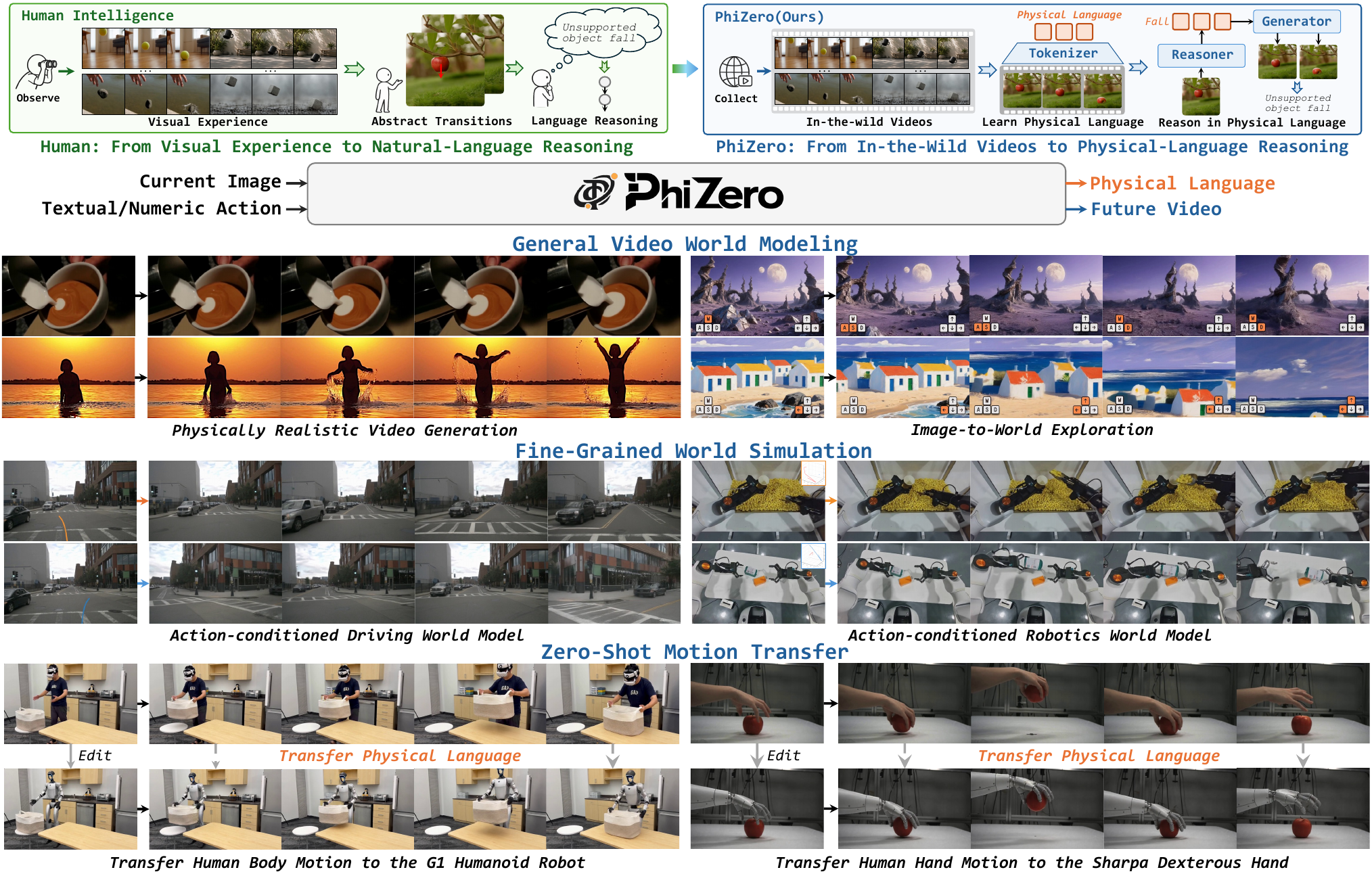}
    \vspace{-1.5em}
    \caption{We introduce \textbf{\Methodname{}}, a physical world model that learns a compact discrete \emph{physical language} from in-the-wild videos. It reasons world evolution in this space and renders the inferred transitions into videos, enabling physically realistic and interactive world modeling, fine-grained action-conditioned simulation, and zero-shot motion transfer.}
    \label{fig:teaser}
\end{figure}

\begin{abstract}

We introduce \Methodname{}, a physical world model built around \emph{physical language}, a compact discrete representation of world-state transitions. Existing physical world models typically predict future videos directly in pixel space, leaving the underlying world dynamics implicit within high-dimensional visual predictors. Motivated by humans' ability to abstract predictive structure from visual experience and organize it in natural language for explicit reasoning, we learn physical language from in-the-wild videos through self-supervision and use it to explicitly reason about how the physical world evolves. Accordingly, \Methodname{} adopts a reason-then-render paradigm: it first infers future world evolution as a physical-language sequence and then renders the inferred transitions into videos. Extensive experiments across generation and understanding benchmarks validate the ability of \Methodname{} to model physically coherent world evolution. We further show its potential for realistic and interactive world modeling, fine-grained action-conditioned simulation, and zero-shot motion transfer.

\end{abstract}

\section{Introduction}
\label{sec:introduction}

\begin{flushright}
    \begin{minipage}{0.58\linewidth}
        \small
        \raggedleft
        \itshape ``You see, but you do not observe.''\\[0.35em]
        \normalfont --- Sherlock Holmes, \emph{A Scandal in Bohemia}
    \end{minipage}
\end{flushright}
\vspace{0.5em}

Video world models can generate future videos with high visual fidelity and are therefore increasingly regarded as promising simulators of the physical world~\citep{openai2025sora2,googledeepmind2026veo31}. For such models to support Physical AI, their central objective must go beyond visual fidelity toward learning how the physical world evolves. However, mainstream approaches are trained primarily through direct prediction in pixel space, leaving the underlying dynamics implicit within high-dimensional visual representations and often resulting in physically inconsistent outcomes.

We draw inspiration from human intelligence. Rather than memorizing individual visual outcomes, humans abstract patterns from visual experience into generalizable knowledge about how the world evolves. Natural language serves as a primary medium for organizing and expressing such knowledge, providing a symbolic space in which humans can explicitly reason. The success of language models across digital domains further demonstrates the scalability of language as a substrate for learning and reasoning. However, when the modeling target shifts from the digital world to the physical world, natural language is often too coarse to faithfully represent the complex state transitions observed in visual experience. We therefore ask: Can we move beyond natural language and learn a finer-grained representation of physical-world transitions that enables explicit reasoning?

To this end, we propose \textbf{\methodname{}}, a physical world model built around \emph{physical language}\physicalscopefootnote. This language is learned from unlabeled in-the-wild videos through self-supervised learning and captures state-transition patterns across diverse visual experiences (Fig.~\ref{fig:teaser}). Once learned, physical language provides an intermediate representation for inferring world evolution before rendering it into video. This reason-then-render paradigm separates dynamics inference from pixel-level synthesis, making world evolution an explicit reasoning target rather than a direct pixel-space prediction.

Learning a physical language that captures how the world evolves requires disentangling world dynamics from visual appearance. We therefore develop a Physical Language Tokenizer that learns this representation through self-supervised video reconstruction. Given a video, the tokenizer encodes its state transitions into a discrete physical-language sequence, which, together with the first frame, conditions a pretrained diffusion decoder to reconstruct the video. The decoder's pretrained generative prior recovers fine-grained visual details, while the first frame anchors static scene appearance, allowing the discrete bottleneck to focus on state changes rather than redundant appearance information. We further introduce a Physical Language Reasoner initialized from a pretrained VLM. Leveraging the VLM's visual-semantic knowledge and commonsense priors, the reasoner autoregressively predicts a physical-language sequence from the current visual state and action intent. The predicted sequence is then rendered into video by the diffusion decoder.

Extensive experiments validate \Methodname{} across video-generation benchmarks assessing physical fidelity and video-understanding benchmarks assessing physical plausibility. We further show that physical language provides a general-purpose interface for representing, controlling, and transferring state transitions in the physical world, supporting interactive rollouts under sequential control inputs, fine-grained action-conditioned world modeling, and zero-shot motion transfer (Fig.~\ref{fig:teaser}). Together, these results position \Methodname{} as a promising framework for controllable and transferable physical-world modeling, with broad potential for Physical AI.

Our main contributions are summarized as follows:
\begin{itemize}
    \item We introduce \emph{physical language}, a compact discrete representation of state-transition patterns that can be learned at scale from in-the-wild videos through self-supervised learning.
    
    \item We develop \Methodname{}, a physical world model built around physical language. \Methodname{} adopts a reason-then-render paradigm, first inferring future world evolution in physical-language space and then rendering the inferred transitions into videos.

    \item Extensive experiments validate \Methodname{} across both physical generation and understanding. We further demonstrate its potential for interactive rollouts, action-conditioned world simulation, and zero-shot motion transfer.
\end{itemize}

\section{Related Work}
\label{sec:related_work}

\subsection{Physical World Models}

Recent work has increasingly developed physical world models by scaling up video generation and prediction to simulate future visual states under language or action conditions. \citet{yang2024unisim} learns interactive simulators from heterogeneous real-world data, while \citet{wang2024worlddreamer,xiang2024pandora} model world evolution through discrete visual prediction and language-conditioned video generation. \citet{agarwal2025cosmos,ali2025world,agarwal2026cosmos3} further advance this paradigm through pretrained world foundation models, unified multimodal conditioning, and omnimodal modeling for Physical AI. Other systems improve long-horizon simulation through autoregressive denoising or latent-state prediction followed by video rendering \citep{zhu2025astra,xiang2025pan}. These approaches primarily represent world evolution in pixel spaces. In contrast, \methodname{} explicitly represents state transitions as a compact discrete physical language, reasons about world evolution in this transition space, and then renders the inferred evolution into video.

\subsection{Learning World Dynamics from Unlabeled Videos}

A broad line of work has investigated how predictive and actionable representations of world dynamics can be learned from unlabeled videos. \citet{bruce2024genie,gao2025adaworld} discover controllable latent action spaces from observed frame-to-frame changes in Internet videos. In robotics and autonomous driving, \citet{schmidt2024LAPO,ye2025LAPA,chen2025moto,bu2025univla,shang2026dynvla} learn latent actions and transfer the resulting motion priors to downstream control tasks. \citet{wu2023ContextWM,wu2024ivideogpt} learn latent dynamics that capture temporal changes transferable across scenes. \citet{bardes2023V-jepa,assran2025V-jepa2} predict masked spatiotemporal regions in representation space, demonstrating that large-scale video pretraining can yield representations useful for understanding and planning in the physical world. \citet{ren2025videoworld,ren2026videoworld2} further investigate whether planning capabilities can be acquired directly from task-oriented videos. Unlike these approaches, we learn a compact physical language that captures state-transition patterns in the physical world, rather than latent actions tied to specific domains or tasks.

\section{Method}
\label{sec:method}

\begin{figure}[h]
    \centering
    \includegraphics[width=0.9\linewidth]{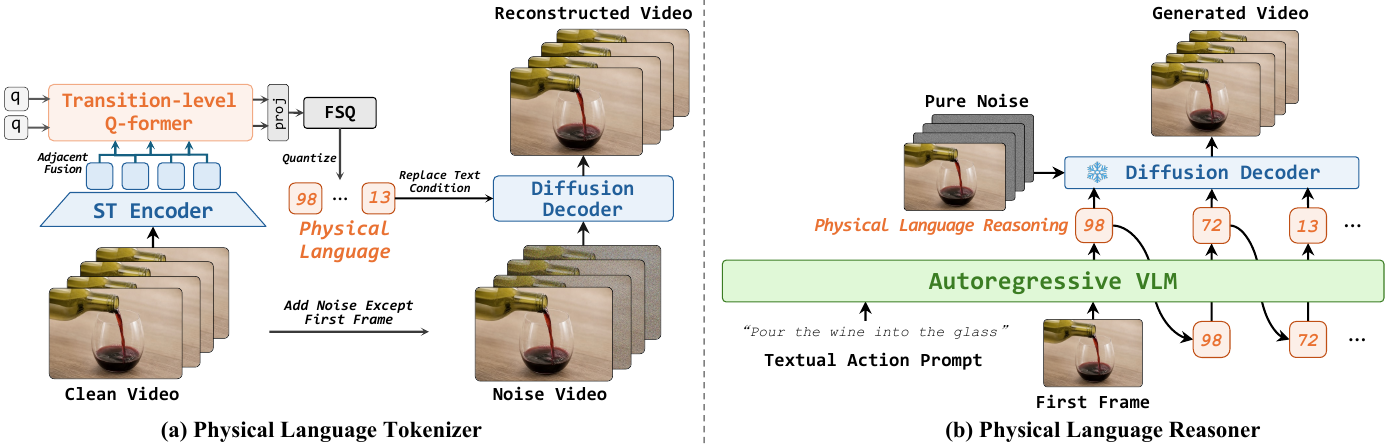}
    \vspace{-0.3em}
    \caption{\textbf{The overall pipeline of \methodname{}.}
    (a) The Physical Language Tokenizer uses a transition-level Q-Former to compress video state transitions into a compact discrete physical language. A diffusion decoder reconstructs the video conditioned on the physical language and the first frame.
    (b) The Physical Language Reasoner uses an autoregressive VLM to predict a physical-language sequence from the first frame and a textual action intent. The trained diffusion decoder then renders the inferred state transition into a future video.}
    \label{fig:pipeline}
\end{figure}

As shown in Fig.~\ref{fig:pipeline}, \methodname{} comprises two complementary components. A Physical Language Tokenizer compresses the state transitions in a video into a discrete physical-language sequence, while a diffusion decoder learns to render the future video conditioned on the first frame and the extracted physical language (Sec.~\ref{sec:physical_tokenizer}). A Physical Language Reasoner then predicts the corresponding physical-language sequence from the first frame and a textual action intent (Sec.~\ref{sec:physical_language_reasoner}).

\subsection{Problem Formulation}
\label{sec:problem_formulation}

Formally, let $\mathbf{V}$ denote the future video, $I^0$ the first frame representing the current world state, $c$ the textual action intent, and $\mathbf{z}$ the discrete physical language describing the transition from $I^0$ to $\mathbf{V}$. Physical language serves as a compact intermediate representation of state transitions that captures shared patterns in how the world evolves. We formulate future prediction as the joint modeling of the physical-language sequence and the future video, which factorizes into physical-language reasoning and video rendering:
\begin{equation}
\underbrace{
p_{\theta,\psi}
\left(
\mathbf{V},\mathbf{z}
\mid
I^0,c
\right)
}_{\text{\Methodname{}}}
=
\underbrace{
p_{\theta}
\left(
\mathbf{z}
\mid
I^0,c
\right)
}_{\substack{\text{Physical-language}\\\text{Reasoning}}}
\underbrace{
p_{\psi}
\left(
\mathbf{V}
\mid
I^0,\mathbf{z}
\right)
}_{\substack{\text{Future-video}\\\text{Rendering}}}.
\label{eq:reason_then_render}
\end{equation}
The Physical Language Reasoner infers how the current state should evolve in the physical-language space, after which the diffusion decoder renders the inferred transition into a future video. This reason-then-render decomposition separates state-transition reasoning from pixel-level synthesis.

\subsection{Physical Language Tokenizer}
\label{sec:physical_tokenizer}

\paragraph{Encoder with Transition-level Q-Former}

Given a video $\mathbf{V} \in \mathbb{R}^{B \times 3 \times T \times H \times W}$, we first employ a spatiotemporal encoder to obtain latent features $\mathbf{x} \in \mathbb{R}^{B \times C \times t \times h \times w}$. Rather than extracting a global representation from the entire video, we explicitly model the transition between each pair of adjacent latent states. This introduces a local temporal inductive bias that reduces the complexity of each compressed transition while preserving the temporal ordering of the full sequence. Let $\mathbf{x}^i \in \mathbb{R}^{B \times C \times h \times w}$ denote the latent state at time step $i$. For each adjacent pair $(\mathbf{x}^i, \mathbf{x}^{i+1})$, we use a shared Q-Former~\citep{li2023blip} to extract a transition representation:
\begin{equation}
\mathbf{q}_i
=
\operatorname{QFormer}
\left(
\mathbf{Q};
\mathbf{x}^i,
\mathbf{x}^{i+1}
\right),
\end{equation}
where $\mathbf{Q} \in \mathbb{R}^{M \times D_q}$ contains $M$ shared learnable transition queries. By jointly attending to the two adjacent latent states, the Q-Former produces $M$ transition features $\mathbf{q}_i \in \mathbb{R}^{B \times M \times D_q}$ for each temporal interval. We concatenate the features from all adjacent intervals in temporal order to obtain $\mathbf{q} \in \mathbb{R}^{B \times N \times D_q}$, where $N=(t-1)M$. We then discretize the transition features using finite scalar quantization (FSQ)~\citep{mentzer2024finite}. Specifically, the features are projected into a low-dimensional quantization space and quantized as
\begin{equation}
\mathbf{z}
=
\operatorname{FSQ}
\left(
\operatorname{Proj}_{\mathrm{down}}(\mathbf{q})
\right),
\end{equation}
where $\mathbf{z}$ denotes the resulting discrete physical-language sequence. FSQ constructs its vocabulary as the Cartesian product of scalar quantization levels and therefore requires no separately learned codebook. Finally, the quantized representation $\mathbf{z}$ is projected to the hidden dimension $d$ of the diffusion transformer, yielding the physical-language context $\mathbf{P}_c \in \mathbb{R}^{B \times N \times d}$.

\paragraph{Diffusion-prior Decoder}

A limited-capacity physical-language bottleneck should focus on representing state transitions rather than encoding all fine-grained appearance details required for video reconstruction. We therefore employ a pretrained video diffusion model~\citep{wan2025wan} as the decoder, whose strong generative prior can recover realistic appearance and visual details from a compact representation of state transitions. To preserve this pretrained generative prior, we retain the original model architecture and replace only its text condition with the physical-language context $\mathbf{P}_c$. We further provide the first frame $I^0$ as a clean visual condition and treat it as the source of static appearance. By supplying appearance information directly through the first frame, the discrete bottleneck is encouraged to encode future state changes rather than redundant static visual details. Conditioned on the physical-language context $\mathbf{P}_c$, the first frame $I^0$, the clean latent $\mathbf{x}_0$, and the diffusion timestep $\tau\sim\mathcal{U}(0,1)$, the decoder reconstructs the target video using the standard flow-matching~\citep{liu2023flow} objective:
\begin{equation}
\mathcal{L}_{\mathrm{FM}}
=
\mathbb{E}_{\mathbf{x},\,\boldsymbol{\epsilon},\,\tau}
\left[
\left\|
v_\psi(\mathbf{x}_\tau,\tau;\,I^0,\mathbf{P}_c)
-
\left(
\boldsymbol{\epsilon}-\mathbf{x}_0
\right)
\right\|_2^2
\right],
\quad
\mathbf{x}_{\tau}
=
(1-\tau)\mathbf{x}_0
+
\tau\boldsymbol{\epsilon},
\quad
\boldsymbol{\epsilon}\sim\mathcal{N}(\mathbf{0},\mathbf{I})
\end{equation}

\begin{figure}[t]
    \centering
    \includegraphics[width=0.9\linewidth]{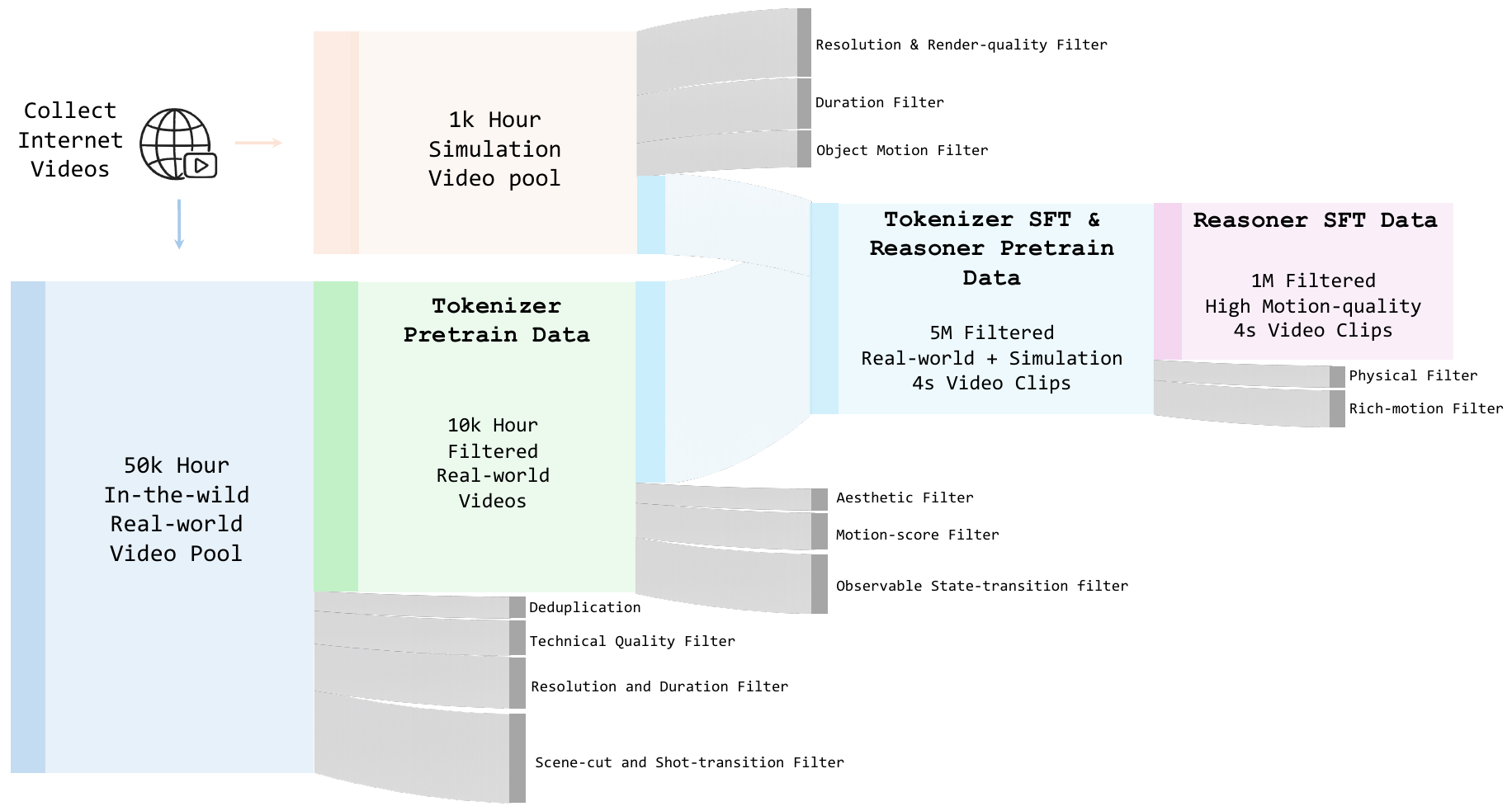}
    \vspace{-0.5em}
    \caption{\textbf{Hierarchical data curation pipeline.} We collect Internet videos to construct a 50K-hour pool of in-the-wild real-world videos and a 1K-hour pool of simulation videos. Progressive filtering yields 10K hours for tokenizer pretraining, 5M four-second clips for tokenizer SFT and reasoner pretraining, and 1M motion-rich four-second clips for reasoner SFT.}
    \label{fig:data}
\end{figure}

\paragraph{Pure-noise Warm-up}

The pretrained diffusion decoder may initially ignore the newly introduced physical-language context by relying on partially corrupted target information and its existing denoising prior. To prevent this shortcut, we introduce a pure-noise warm-up stage in which all future-frame latents are initialized from pure noise. The decoder must therefore rely on the physical-language context and the first frame to reconstruct the future video. After warm-up, we restore the standard flow-matching noise schedule. This strategy mitigates the denoising shortcut and provides stronger supervision for learning informative physical-language representations.

\paragraph{Data Curation Pipeline}
We train the Physical Language Tokenizer in two stages: large-scale pretraining followed by targeted SFT. As shown in Fig.~\ref{fig:data}, for pretraining, we begin with a real-world video pool containing approximately 50K hours of footage. We remove duplicates and filter out clips with technical defects, including compression artifacts, corrupted frames, and watermarks, as well as those that fail the resolution or duration requirements or contain abrupt shot transitions. This process yields approximately 10K hours of unlabeled videos for tokenizer pretraining. For the subsequent SFT stage, we apply a stricter second-pass filtering procedure based on aesthetic quality, motion magnitude, and the observability of state transitions as assessed by a VLM judge. In parallel, we curate approximately 1K hours of simulated videos by removing samples with low rendering quality, invalid duration, or insufficient object motion. Combining the resulting real-world and simulation data produces approximately 5M four-second video clips for tokenizer SFT. Additional data details are provided in the Appendix~\ref{app:physical_language_tokenizer}.

\paragraph{Curriculum Training Recipe}
We couple this two-stage training scheme with a joint temporal and spatial curriculum. During pretraining, we process videos at a resolution of $256\times448$ and progressively increase the clip duration from 1 second to 2 seconds and finally to 4 seconds. This allows the tokenizer to first learn local state changes before modeling longer-range world evolution. We then increase the reconstruction resolution to $512\times896$ and perform SFT on the curated 5M-clip corpus. Finally, we freeze the tokenizer and refine only the diffusion decoder to further improve reconstruction quality. Additional training details are provided in Appendix~\ref{app:physical_language_tokenizer}.

\subsection{Physical Language Reasoner}
\label{sec:physical_language_reasoner}

\paragraph{Physical Language Reasoning with a Pretrained VLM}

Predicting future state transitions requires interpreting a textual action intent in the context of the current visual state and inferring how the world should evolve. Pretrained VLMs encode rich visual semantics and commonsense world knowledge from large-scale image-text data, providing a strong prior for this task. We thus initialize the Physical Language Reasoner from a pretrained VLM~\citep{bai2025qwen3vl}. To adapt it to the physical-language space learned by the Physical Language Tokenizer, we extend its vocabulary with a distinct atomic symbol for each FSQ index. Given a first frame $I^0$ and a text prompt $c$, the Physical Language Reasoner autoregressively predicts a length-$N$ physical-language sequence over the FSQ-induced vocabulary of size $K$. Its conditional distribution is factorized in temporal order as
\begin{equation}
p_{\theta}(\mathbf{z}\mid I^0,c)
=
\prod_{j=1}^{N}
p_{\theta}(z_j\mid I^0,c,z_{<j}).
\end{equation}
We generate supervision offline by encoding each training video with the frozen Physical Language Tokenizer, $\mathbf{z}=\mathcal{T}_{\phi}(\mathbf{V})$. Under teacher forcing, the Physical Language Reasoner is optimized with the autoregressive cross-entropy objective
\begin{equation}
\mathcal{L}_{\mathrm{VLM}}
=
-\sum_{j=1}^{N}
\log p_{\theta}(z_j\mid I^0,c,z_{<j}).
\end{equation}

\paragraph{Data Curation Pipeline}

We construct the training data for the Physical Language Reasoner in two stages of increasing specialization. As shown in Fig.~\ref{fig:data}, for continued pretraining, we reuse the approximately 5M curated four-second real-world and simulation clips used for Physical Language Tokenizer SFT. For each clip, a VLM-generated caption and the first frame serve as inputs, while the frozen Physical Language Tokenizer provides the target physical-language sequence. To prevent the textual condition from revealing the outcome, as shown in Table~\ref{tab:caption_prompt}, we prompt the VLM to summarize only the high-level initiating action or interaction in each clip, rather than narrating fine-grained transition details. To construct the subsequent SFT corpus, we further filter this general dataset using a VLM-based rich-motion filter and a physical filter, retaining clips that exhibit salient state changes and physically informative interactions. Together with curated simulator-generated samples, this process yields approximately 1M motion-rich and physically informative four-second clips for reasoner SFT. Additional data details are provided in the Appendix~\ref{app:physical_language_reasoner}.

\paragraph{Two-stage Training Recipe}
We train the Physical Language Reasoner in two stages with progressively specialized data distributions. In Stage 1, we initialize the model from a pretrained VLM~\citep{bai2025qwen3vl} and perform continued pretraining on the 5M-clip general corpus. This stage adapts the VLM to autoregressive physical-language prediction and establishes the correspondence among textual intent, the current visual state, and the resulting state transition. In Stage 2, we perform SFT on a corpus of approximately 1M motion-rich and physically informative clips. By concentrating training on interactions with salient motion and physically meaningful consequences, this stage improves the physical plausibility and precision of the inferred state transitions while preserving the broad capability for modeling world transitions acquired during continued pretraining. Additional training details are provided in the Appendix~\ref{app:physical_language_reasoner}.

\begin{table}[t]
    \centering
    \begin{minipage}[t]{0.495\linewidth}
        \centering
        \begingroup
        \scriptsize
        \setlength{\abovecaptionskip}{2pt}
        \compacttablecaption{\textbf{Physical outcome fidelity on Physics-IQ Verified.}\label{tab:physics_iq_verified}}
        \endgroup
        \scriptsize
        \setlength{\tabcolsep}{2.2pt}
        \renewcommand{\arraystretch}{1.00}
        \resizebox{0.98\linewidth}{!}{%
        \begin{tabular}{lccc>{\columncolor{totalmetricgray}}c}
            \toprule
            \textbf{Model} & \textbf{S-IoU $\uparrow$} & \textbf{ST-IoU $\uparrow$} & \textbf{WS-IoU $\uparrow$}
            & \cellcolor{white}\textbf{IQ-Score $\uparrow$} \\
            \midrule
            Wan2.2-5B~\citep{wan2025wan} & 24.7 & 22.6 & 13.3
            & 21.2 \\
            Sora 2~\citep{openai2025sora2} & 37.3 & \underline{27.0} & 26.9
            & 26.5 \\
            Cosmos3-Nano~\citep{agarwal2026cosmos3} & 40.4 & 22.0 & 24.6
            & 29.1 \\
            Wan2.2-14B~\citep{wan2025wan} & {51.1} & 20.5 & 28.5
            & 32.2 \\
            Hunyuan-Video~\citep{kong2024hunyuanvideo} & 47.1 & {26.9} & \underline{29.7}
            & 33.4 \\
            Grok-Video~\citep{xai2026grokvideo} & \underline{52.7} & 21.4 & \textbf{35.7}
            & 34.8 \\
            Cosmos3-Super~\citep{agarwal2026cosmos3} & -- & -- & --
            & \underline{39.5} \\
            \midrule
            \textbf{\methodname{} (Ours)} & \textbf{58.2} & \textbf{36.8} & 27.6
            & \textbf{41.2} \\
            \bottomrule
        \end{tabular}%
        }
    \end{minipage}\hfill
     \begin{minipage}[t]{0.495\linewidth}
          \centering
          \begingroup
          \scriptsize
          \setlength{\abovecaptionskip}{2pt}
          \compacttablecaption{\textbf{Physical-law adherence on PhyGround.}\label{tab:phyground}}
          \endgroup
          \scriptsize
          \setlength{\tabcolsep}{2.2pt}
          \renewcommand{\arraystretch}{1.06}
          \resizebox{0.98\linewidth}{!}{%
          \begin{tabular}{lcc>{\columncolor{totalmetricgray}}c}
              \toprule
              \textbf{Model} & \textbf{General Quality $\uparrow$} & \textbf{Physics Score $\uparrow$} & \cellcolor{white}\textbf{Overall $\uparrow$} \\
              \midrule
              LTX-Video-19B~\citep{hacohen2026ltx2} & 2.56 & 2.54 & 2.55 \\
              LTX-Video-22B~\citep{hacohen2026ltx2} & 2.59 & 2.56 & 2.57 \\
              Wan2.2-5B~\citep{wan2025wan} & 2.67 & 2.65 & 2.66 \\
              Cosmos2.5-2B~\citep{gu2025cosmospredict} & 2.79 & 2.70 & 2.74 \\
              OmniWeaving~\citep{pan2026omniweaving} & 2.89 & 2.78 & 2.84 \\
              Veo3.1~\citep{googledeepmind2026veo31} & \textbf{3.01} & 2.85 & 2.93 \\
              Wan2.2-14B~\citep{wan2025wan} & \underline{3.00} & \underline{2.90} & \underline{2.95} \\
              \midrule
              \textbf{\methodname{} (Ours)} & 2.93 & \textbf{3.01} & \textbf{2.97} \\
              \bottomrule
          \end{tabular}%
          }
      \end{minipage}
\end{table}

\begin{table}[t]
    \centering
    \begin{minipage}[t]{0.495\linewidth}
        \centering
        \begingroup
        \scriptsize
        \setlength{\abovecaptionskip}{2pt}
        \compacttablecaption{\textbf{General world modeling on WorldModelBench.}\label{tab:worldmodelbench}}
        \endgroup
        \vspace{2pt}
        \scriptsize
        \setlength{\tabcolsep}{3.0pt}
        \renewcommand{\arraystretch}{1.07}
        \resizebox{0.98\linewidth}{!}{%
        \begin{tabular}{lcc>{\columncolor{totalmetricgray}}c}
            \toprule
            \textbf{Model} & \textbf{Physics Adherence $\uparrow$} & \textbf{Common Sense $\uparrow$} & \cellcolor{white}\textbf{Total $\uparrow$} \\
            \midrule
            Pandora~\citep{xiang2024pandora} & 4.05 & 0.95 & 6.57 \\
            OpenSora-Plan~\citep{lin2024opensoraplan} & 3.85 & 1.01 & 6.62 \\
            CogVideoX~\citep{yang2025cogvideox} & 3.88 & 0.99 & 6.75 \\
            Mochi~\citep{genmoaiblog2024Mochi} & 4.14 & 1.26 & 7.62 \\
            Luma~\citep{lumaai2024luma} & 4.13 & 1.57 & 7.72 \\
            Wan2.2-5B~\citep{wan2025wan} & \underline{4.51} & 1.41 & 7.98 \\
            Runway~\citep{runway2024runway} & 4.27 & \underline{1.65} & \underline{8.08} \\
            \midrule
            \textbf{\methodname{} (Ours)} & \textbf{4.88} & \textbf{1.71} & \textbf{8.19} \\
            \bottomrule
        \end{tabular}%
        }
    \end{minipage}\hfill
    \begin{minipage}[t]{0.495\linewidth}
        \centering
        \begingroup
        \scriptsize
        \setlength{\abovecaptionskip}{2pt}
        \compacttablecaption{\textbf{Intuitive physics understanding on IntPhys2.}\label{tab:intphys2}}
        \endgroup
        \scriptsize
        \setlength{\tabcolsep}{3.0pt}
        \renewcommand{\arraystretch}{1.06}
        \resizebox{0.98\linewidth}{!}{%
        \begin{tabular}{lccc>{\columncolor{totalmetricgray}}c}
            \toprule
            \textbf{Model} & \textbf{Easy $\uparrow$} & \textbf{Medium $\uparrow$} & \textbf{Hard $\uparrow$} & \cellcolor{white}\textbf{Overall $\uparrow$} \\
            \midrule
            Cosmos-4B~\citep{agarwal2025cosmos} & 46.00 & 52.00 & 48.05 & 49.41 \\
            Qwen2.5-VL~\citep{Qwen2.5-VL} & 50.96 & 53.25 & 51.49 & 52.27 \\
            Gemini-1.5 Pro~\citep{geminiteam2023gemini} & {58.65} & 53.00 & 52.67 & 52.27 \\
            GPT-4o~\citep{achiam2023gpt4} & 57.69 & 54.75 & 54.17 & 53.75 \\
            VideoMAEv2~\citep{wang2023videomaev2} & 46.00 & \underline{58.50} & 52.73 & 53.75 \\
            V-JEPA~\citep{bardes2023V-jepa} & 52.00 & 53.00 & \textbf{57.42} & 53.75 \\
            Gemini-2.5 Flash~\citep{geminiteam2023gemini} & \textbf{64.42} & 56.75 & \underline{54.46} & \underline{55.63} \\
            \midrule
            \textbf{\methodname{} (Ours)} & \underline{60.98} & \textbf{60.50} & {52.38} & \textbf{56.34} \\
            \bottomrule
        \end{tabular}%
        }
    \end{minipage}
\end{table}

\begin{table}[t]
    \centering
    \begin{minipage}[t]{0.495\linewidth}
        \centering
        \begingroup
        \scriptsize
        \setlength{\abovecaptionskip}{2pt}
        \compacttablecaption{\textbf{Physical-plausibility discrimination on LikePhys.}\label{tab:likephys}}
        \endgroup
        \scriptsize
        \setlength{\tabcolsep}{2.3pt}
        \renewcommand{\arraystretch}{1.06}
        \resizebox{0.98\linewidth}{!}{%
        \begin{tabular}{lccc>{\columncolor{totalmetricgray}}c}
            \toprule
            \textbf{Model}
            & \textbf{Rigid $\downarrow$}
            & \textbf{Fluid $\downarrow$}
            & \textbf{Optical $\downarrow$}
            & \cellcolor{white}\textbf{Avg. Error $\downarrow$} \\
            \midrule
            AnimateDiff-SDXL~\citep{guo2023animatediff} & 61.44 & 53.33 & 37.65 & 56.0 \\ 
            ZeroScope~\citep{wang2023zeroscope} & 57.32 & 49.23 & 41.00 & 53.3 \\ 
            ModelScope~\citep{wang2023zeroscope} & 59.46 & 47.23 & 35.65 & 52.9 \\ 
            Mochi~\citep{genmoaiblog2024Mochi} & 54.22 & 47.33 & 49.15 & 51.9 \\ 
            CogVideoX-5B~\citep{yang2025cogvideox} & 59.00 & 43.10 & \underline{26.65} & 49.8 \\ 
            CogVideoX-2B~\citep{yang2025cogvideox} & 56.98 & \underline{42.00} & \textbf{25.15} & 48.2 \\ 
            Wan2.1-1.3B~\citep{wan2025wan} & 44.66 & 57.10 & 41.65 & 48.0 \\ 
            LTX-Video-2B~\citep{hacohen2026ltx2} & 43.44 & \textbf{33.43} & 61.50 & 44.7 \\ 
            CogVideoX1.5-5B~\citep{yang2025cogvideox} & 44.14 & 47.90 & 42.65 & {43.8} \\ 
            Wan2.1-14B~\citep{wan2025wan} & 43.34 & 45.00 & 38.35 & {43.8} \\ 
            Hunyuan-Video~\citep{kong2024hunyuanvideo} & \underline{36.34} & 48.67 & 41.15 & \underline{43.6} \\ 
            \midrule
            \textbf{\methodname{} (Ours)} & \textbf{29.14} & 53.15 & 37.50 & \textbf{41.7} \\ 
            \bottomrule
        \end{tabular}%
        }
    \end{minipage}\hfill
    \begin{minipage}[t]{0.495\linewidth}
        \centering
        \begingroup
        \scriptsize
        \setlength{\abovecaptionskip}{2pt}
        \compacttablecaption{\textbf{Real-world causal understanding on YoCausal.}\label{tab:yocausal}}
        \endgroup
        \vspace{1.5pt}
        \scriptsize
        \setlength{\tabcolsep}{2.0pt}
        \renewcommand{\arraystretch}{0.98}
        \resizebox{0.98\linewidth}{!}{%
        \begin{tabular}{lcc>{\columncolor{totalmetricgray}}c}
            \toprule
            \textbf{Model}
            & \textbf{RSI (\%) $\uparrow$}
            & \textbf{CCI (\%) $\uparrow$}
            & \cellcolor{white}\textbf{Agg. Rank $\downarrow$} \\
            \midrule
            CogVideoX-2B~\citep{yang2025cogvideox}
            & 41.50 & 0.93 & 10.0 \\ 
            Wan2.2-5B~\citep{wan2025wan}
            & 51.91 & -2.12 & 9.0 \\ 
            Hunyuan-Video~\citep{kong2024hunyuanvideo}
            & 52.05 & -0.29 & 8.0 \\ 
            Wan2.1-1.3B~\citep{wan2025wan}
            & 45.51 & 5.36 & 7.5 \\ 
            Mochi~\citep{genmoaiblog2024Mochi}
            & 49.12 & 3.85 & 8.0 \\ 
            CogVideoX1.5-5B~\citep{yang2025cogvideox}
            & 46.83 & 4.85 & 8.0 \\ 
            CogVideoX-5B~\citep{yang2025cogvideox}
            & 49.92 & 5.09 & 6.5 \\ 
            LTX-Video-13B~\citep{hacohen2026ltx2}
            & \underline{56.48} & -4.32 & 7.0 \\ 
            LTX-Video-2B~\citep{hacohen2026ltx2}
            & \textbf{58.86} & -0.20 & 5.0 \\ 
            Wan2.1-14B~\citep{wan2025wan}
            & 53.24 & \underline{5.91} & \underline{3.5} \\ 
            Wan2.2-14B~\citep{wan2025wan}
            & 54.19 & 5.51 & \underline{3.5} \\ 
            \midrule
            \textbf{\methodname{} (Ours)}
            & 55.54 & \textbf{6.20} & \textbf{2.0} \\ 
            \bottomrule
        \end{tabular}%
        }
    \end{minipage}
\end{table}

\section{Experiments}
\label{sec:experiments}

\subsection{Implementation Details}
\label{sec:implementation_details}

For the Physical Language Tokenizer, the spatiotemporal encoder follows the Wan2.2 VAE encoder architecture~\citep{wan2025wan} and is initialized from its pretrained weights, while the diffusion decoder is initialized from Wan2.2-5B~\citep{wan2025wan}. We optimize all parameters of the spatiotemporal encoder and transition-level Q-Former, while fine-tuning the diffusion decoder using LoRA~\citep{hu2022lora} with rank 32. We configure FSQ with scalar quantization levels $(8,5,5,5,5,5)$, yielding a vocabulary of 25K discrete symbols, and use 32 learnable queries in the transition-level Q-Former. A 33-frame video is encoded into nine temporal latent states; extracting 32 transition symbols from each adjacent pair therefore produces a physical-language sequence of length $(9-1)\times32=256$. For the Physical Language Reasoner, we initialize the model from the pretrained Qwen3-VL-4B~\citep{bai2025qwen3vl} and extend its vocabulary with one atomic symbol for each FSQ index. Additional training configurations and data details are provided in Appendix~\ref{app:training_details}.

\begin{figure}[t]
    \centering
    \includegraphics[width=\linewidth]{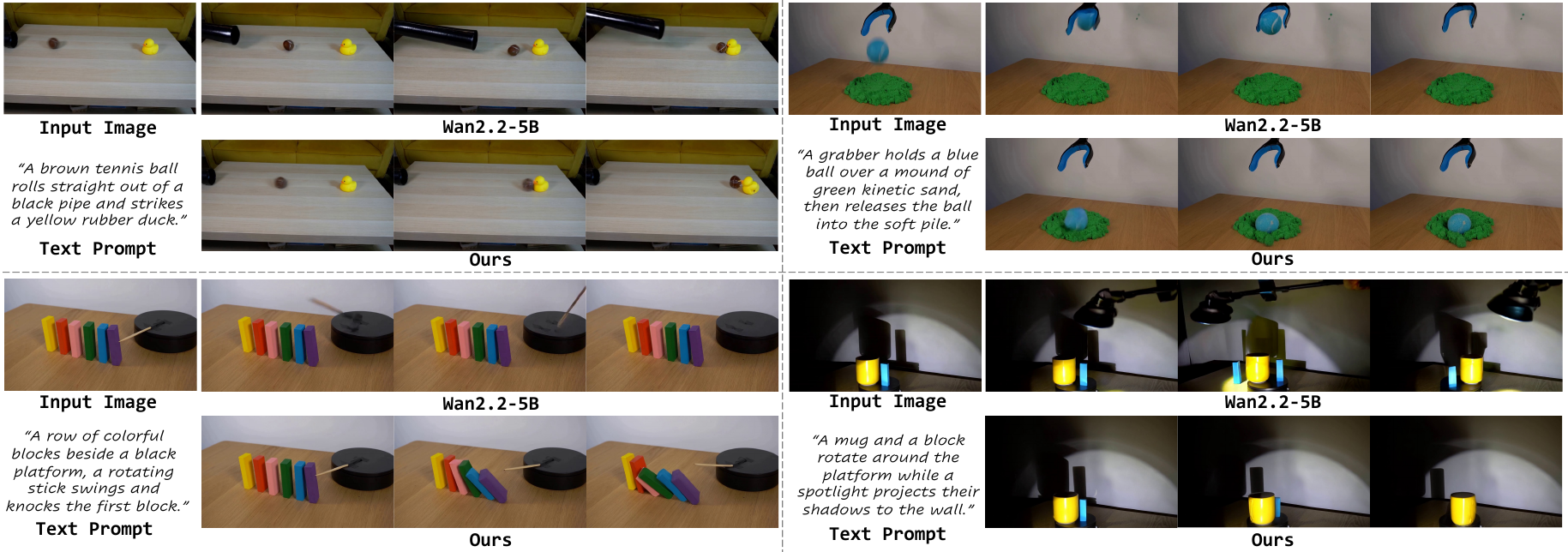}
    \vspace{-1em}
    \caption{\textbf{Qualitative comparison on Physics-IQ Verified.}
    Compared with the Wan2.2-5B baseline, \Methodname{} better captures the physical consequences of interactions, including collision-induced object displacement, gravity-driven deformation, chain reactions, and dynamically changing shadows.}
    \label{fig:qualitative_physics_iq}
\end{figure}

\begin{table}[t]
    \centering
    \begingroup
    \footnotesize
    \setlength{\abovecaptionskip}{2pt}
    \caption{\textbf{Reconstruction performance of the Physical Language Tokenizer.} Results are reported on 500 four-second real-world videos sampled at 8 FPS and evaluated at a resolution of $512 \times 896$.}
    \label{tab:physical_tokenizer_reconstruction_comparison}
    \endgroup
    \footnotesize
    \setlength{\tabcolsep}{5pt}
    \renewcommand{\arraystretch}{1.08}
    \begin{tabular}{lcccc}
        \toprule
        \textbf{Model} & \textbf{Tokens} $\downarrow$ & \textbf{PSNR} $\uparrow$ & \textbf{SSIM} $\uparrow$ & \textbf{LPIPS} $\downarrow$ \\
        \midrule
        \textcolor{gray}{Wan2.2-5B VAE\citep{wan2025wan}} & \textcolor{gray}{44800} & \textcolor{gray}{37.7} & \textcolor{gray}{0.957} & \textcolor{gray}{0.042} \\
        \midrule
        Video-LaVIT~\citep{jin2024videolavit} & 270 & 23.6 & 0.835 & 0.175 \\
        VideoFlexTok~\citep{atanov2026videoflextok} ($k=32$) & 288 & 25.2 & 0.858 & 0.177 \\
        VideoFlexTok~\citep{atanov2026videoflextok} ($k=64$) & 576 & 26.5 & 0.870 & 0.167 \\
        \midrule
        Ours & \textbf{256} & \textbf{28.9} & \textbf{0.903} & \textbf{0.087} \\
        \bottomrule
    \end{tabular}
\end{table}

\subsection{Main Results}
\label{sec:main_results}

\paragraph{Video Generation Results}

We evaluate \methodname{} on three video-generation benchmarks for world models. Physics-IQ Verified~\citep{radsch2026physicsiqverified} measures physical outcome fidelity by comparing generated videos with real reference videos using rule-based metrics. PhyGround~\citep{lin2026phyground} evaluates physical-law adherence with physics-specialized VLM judges, while WorldModelBench~\citep{li2026worldmodelbench} assesses general world-modeling capability in terms of physics adherence and overall video quality. Additional details on the generation benchmarks, evaluation protocols, and metrics are provided in Appendix~\ref{app:physical_generation_evaluation}. As shown in Tables~\ref{tab:physics_iq_verified}, \ref{tab:phyground}, and \ref{tab:worldmodelbench}, \methodname{} achieves strong performance across all three generation benchmarks. It obtains the best IQ-Score on Physics-IQ Verified, the highest Physics Score and Overall score on PhyGround, and the best Physics Adherence and Total scores on WorldModelBench. These results demonstrate that \methodname{} generates state transitions that closely match real-world physical outcomes across diverse open-domain phenomena.

\paragraph{Video Understanding Results}

We further evaluate \methodname{} on three video-understanding benchmarks. IntPhys2~\citep{bordes2025intphys2} evaluates intuitive physics understanding, LikePhys~\citep{yuan2025likephys} evaluates physical-plausibility discrimination in simulated scenes, and YoCausal~\citep{xie2026yocausal} examines temporal and causal understanding in real-world videos. All three benchmarks adopt a pairwise comparison protocol, requiring the model to distinguish a physically or causally valid video from a matched invalid counterpart. For each pair, we encode both videos into physical-language sequences, compute their log-likelihoods under the Physical Language Reasoner, and select the video with the higher likelihood as valid. Additional details on the understanding benchmarks and our likelihood-based evaluation protocol are provided in Appendix~\ref{app:physical_understanding_evaluation}. As shown in Tables~\ref{tab:intphys2}, \ref{tab:likephys}, and \ref{tab:yocausal}, \methodname{} achieves competitive performance across all three understanding benchmarks, demonstrating its ability to identify physical inconsistencies and causal violations.

\paragraph{Qualitative Results}

As shown in Fig.~\ref{fig:qualitative_physics_iq}, we compare \Methodname{} with the Wan2.2-5B baseline~\citep{wan2025wan} on four representative cases from Physics-IQ Verified~\citep{radsch2026physicsiqverified}. Although Wan2.2-5B often produces visually plausible interactions, it fails to generate their expected physical consequences. For example, the tennis ball strikes the rubber duck, but the duck remains unaffected by the collision. In contrast, \Methodname{} generates coherent downstream effects, including collision-induced object displacement, deformation under gravity, chain reactions, and changing shadows.

\paragraph{Physical Language Tokenizer Reconstruction Performance}

We further evaluate the reconstruction performance of the Physical Language Tokenizer on 500 four-second real-world videos sampled at 8 FPS and evaluated at a resolution of $512 \times 896$. We report the number of representation tokens excluding the first-frame condition. As shown in Table~\ref{tab:physical_tokenizer_reconstruction_comparison}, the Wan2.2 VAE achieves high reconstruction quality but requires 44,800 continuous visual tokens. In contrast, our Physical Language Tokenizer represents the future state transitions in each video with only 256 discrete physical-language symbols and achieves the best reconstruction quality among the highly compressed tokenizers. These results demonstrate that physical language can compactly encode state transitions while retaining sufficient information for high-quality video reconstruction.

\subsection{Ablation Studies}
\label{sec:ablation}

\begin{wraptable}{l}{0.48\textwidth}
    \vspace{-0.75\baselineskip}
    \centering
    \begingroup
    \footnotesize
    \setlength{\abovecaptionskip}{2pt}
    \ablationtablecaption{\textbf{Physical Language Tokenizer Ablations}\label{tab:physical_language_tokenizer_design_ablation}}
    \endgroup
    \footnotesize
    \setlength{\tabcolsep}{5pt}
    \renewcommand{\arraystretch}{1.08}
    \resizebox{\linewidth}{!}{%
        \begin{tabular}{lccc}
            \toprule
            \textbf{Model} & \textbf{PSNR} $\uparrow$ & \textbf{SSIM} $\uparrow$ & \textbf{LPIPS} $\downarrow$ \\
            \midrule
            w/o Diffusion Dec & 26.6 & 0.875 & 0.119 \\
            w/o Trans. Q-Former & 28.2 & 0.896 & 0.089 \\
            w/o Pure-noise Warm-up & 27.9 & 0.894 & 0.091 \\
            \midrule
            \textbf{Full} & \textbf{28.9} & \textbf{0.903} & \textbf{0.087} \\
            \bottomrule
        \end{tabular}%
    }
    \vspace{-0.4\baselineskip}
\end{wraptable}

\paragraph{Physical Language Tokenizer Design Ablations}

We ablate the key design components of the Physical Language Tokenizer in Table~\ref{tab:physical_language_tokenizer_design_ablation}. Replacing the pretrained diffusion decoder with a conventional deterministic decoder causes the largest performance degradation, showing that the diffusion prior recovers fine-grained appearance details and allows the compact bottleneck to focus on state transitions. Replacing the transition-level Q-Former with a global Q-Former also degrades reconstruction quality, validating the local temporal inductive bias introduced by explicitly modeling transitions between adjacent latent states. Finally, removing the pure-noise warm-up consistently reduces reconstruction performance, indicating that the warm-up encourages the decoder to rely on the physical-language condition rather than relying primarily on its pretrained denoising prior.

\begin{figure}[t]
    \centering
    \includegraphics[width=1.0\linewidth]{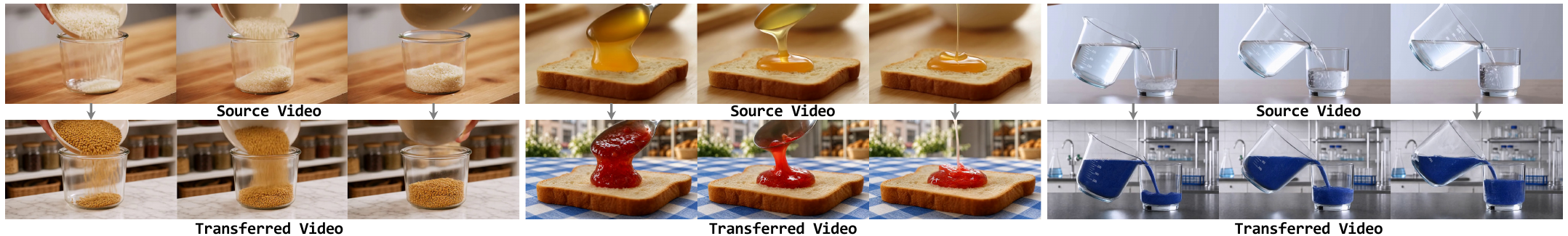}
    \vspace{-1.5em}
    \caption{\textbf{Transferability of Physical Language.}
    We encode each source video's state transition into a physical-language sequence and decode the same sequence conditioned on an edited first frame with a different visual appearance. The transferred videos preserve the source evolution across substantial changes in object and scene appearance.}
    \label{fig:physical_language_transfer}
\end{figure}

\begin{figure}[t]
    \centering
    \begin{minipage}[t]{0.4\linewidth}
        \centering
        \includegraphics[width=\linewidth]{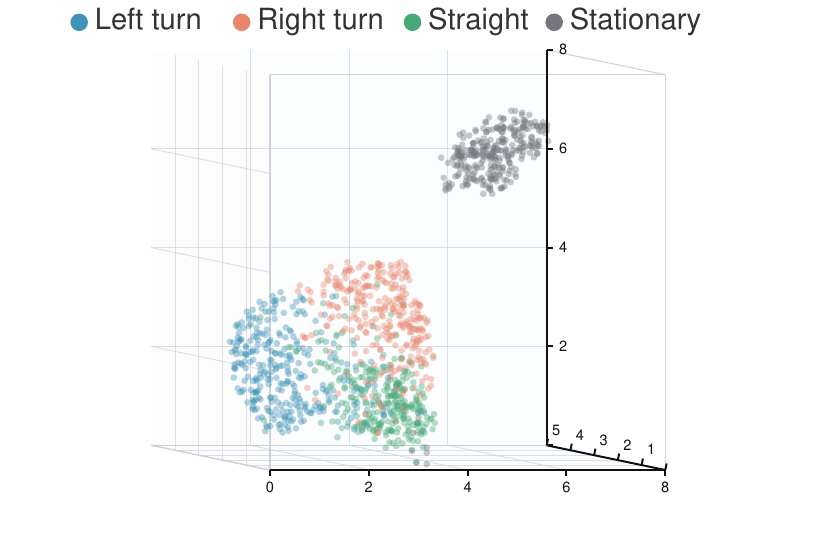}
        \vspace{-0.3\baselineskip}
        {\small (a) Autonomous driving}
    \end{minipage}\hspace{-0.05\linewidth}%
    \begin{minipage}[t]{0.4\linewidth}
        \centering
        \includegraphics[width=\linewidth]{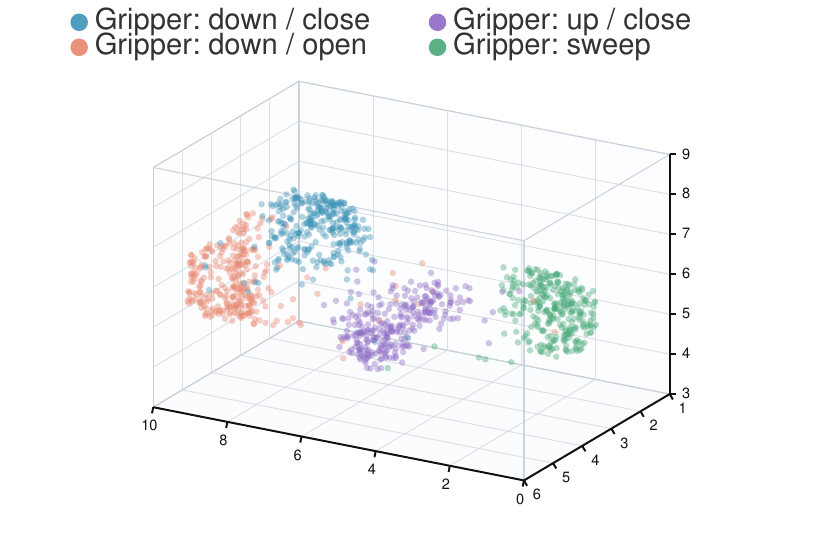}
        \vspace{-0.3\baselineskip}
        {\small (b) Robotic manipulation}
    \end{minipage}
    \vspace{0.3em}
    \caption{\textbf{Semantic structure of physical language.}
    We aggregate transition features into clip-level representations, reduce them to 20 dimensions using PCA, and project them into 3D using UMAP. The resulting embeddings naturally organize according to the underlying transition patterns, forming structured manifolds or clusters across the two domains.}
    \label{fig:physical_language_structure}
\end{figure}

\begin{wraptable}{l}{0.35\textwidth}
    \vspace{-0.8\baselineskip}
    \centering
    \begingroup
    \footnotesize
    \setlength{\abovecaptionskip}{2pt}
    \ablationtablecaption{\textbf{Reasoner Ablations}\label{tab:physical_reasoner_design_ablation}}
    \endgroup
    \footnotesize
    \setlength{\tabcolsep}{5pt}
    \renewcommand{\arraystretch}{1.08}
    \resizebox{\linewidth}{!}{%
        \begin{tabular}{lc}
            \toprule
            \textbf{Method} & \textbf{IQ-Score $\uparrow$} \\
            \midrule
            Wan2.2-5B (Baseline) & 21.2 \\
            + Prompt Enhancement & 26.6 \\
            \midrule
            Ours w/o Simulation Data & 37.7 \\
            Ours w/o Two-stage Training & 39.2 \\
            \midrule
            \textbf{Ours (Full)} & \textbf{41.2} \\
            \bottomrule
        \end{tabular}%
    }
    \vspace{-1\baselineskip}
\end{wraptable}

\paragraph{Physical Language Reasoner Design Ablations}

We evaluate the Physical Language Reasoner on Physics-IQ Verified. As shown in Table~\ref{tab:physical_reasoner_design_ablation}, prompt enhancement improves the Wan2.2-5B baseline, but still remains substantially below \methodname{}, indicating that natural-language reasoning alone is insufficient for modeling fine-grained state transitions. Removing simulation data also reduces performance, confirming that simulator-generated interactions improve the prediction of physically plausible transitions and benefit real-world video generation. This result further highlights the potential to scale physical-language reasoning with simulator-generated data. Finally, replacing two-stage training with joint training degrades performance, showing that dedicated SFT on a curated corpus of motion-rich and physically informative videos helps the model infer more precise and physically plausible state transitions.

\begin{figure}[t]
    \centering
    \includegraphics[width=1.0\linewidth]{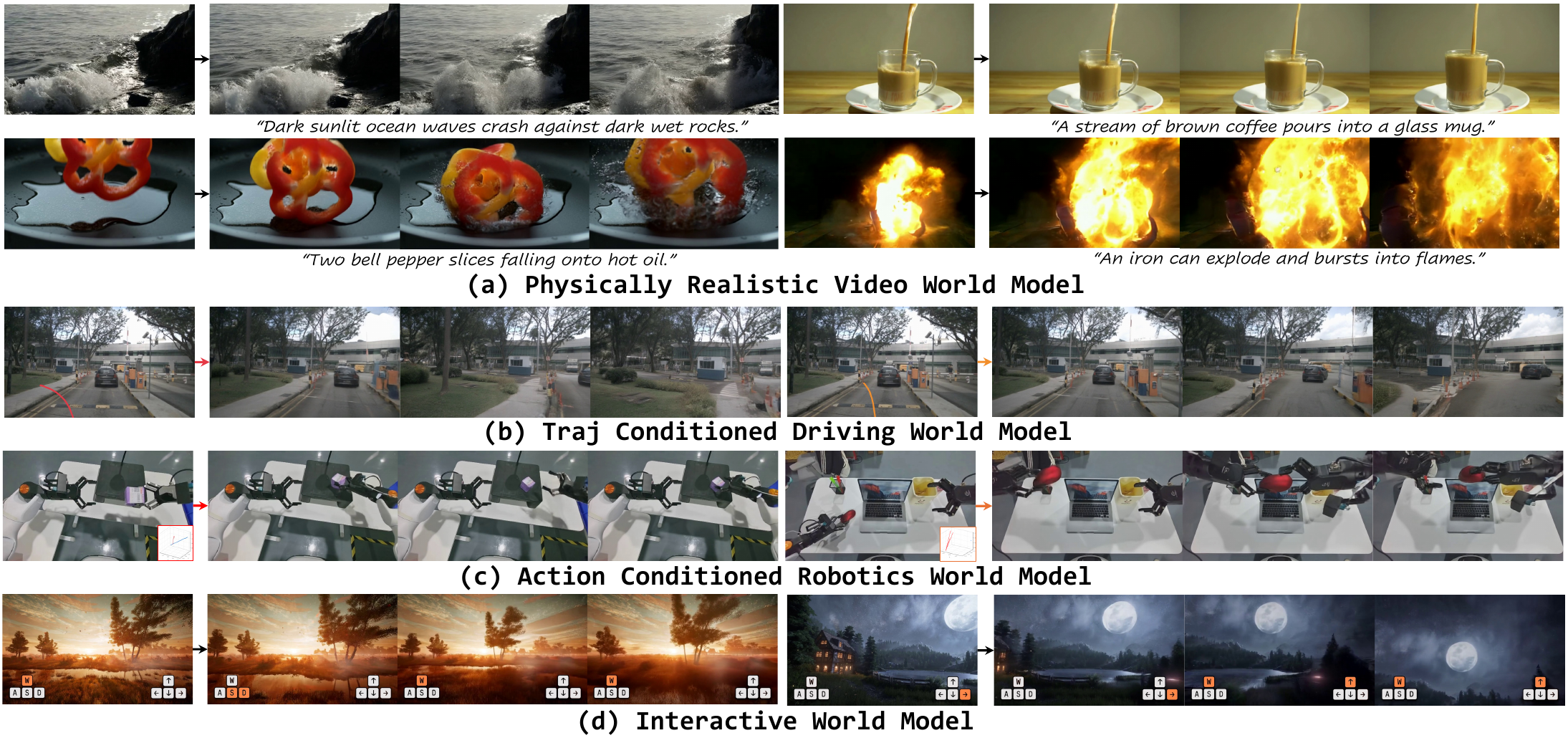}
    \vspace{-1.5em}
    \caption{\textbf{\Methodname{} as a physically realistic, controllable, and interactive video world model.}
    \Methodname{} generates physically coherent world evolution, follows fine-grained action conditions, and supports interactive rollouts under sequential controls.}
    \label{fig:physical_generation}
\end{figure}

\subsection{Analysis of Physical Language}

\paragraph{Transferability of Physical Language}

We investigate whether physical language disentangles state transitions from visual appearance. Given a source video, we encode its state transition with the Physical Language Tokenizer, edit the first frame to specify a different target appearance, and decode the unchanged physical-language sequence conditioned on the edited frame. As shown in Fig.~\ref{fig:physical_language_transfer}, the transferred videos preserve transition patterns across changes in appearance and background, including the pouring, viscous spreading, and liquid flow. These results indicate that physical language captures reusable transition information that can be transferred across visual appearances.

\paragraph{Semantic Structure of Physical Language}

We analyze the semantic structure of physical language in two domains with clearly defined kinematic patterns: autonomous driving and robotic manipulation. We use Physical Language Tokenizers adapted to the corresponding domains and select four categories with 300 samples per category. For autonomous driving, we select left turn, right turn, straight driving, and stationary clips from nuScenes~\citep{caesar2020nuscenes}. For robotic manipulation, we select four gripper transition patterns from AGI-Bot RealRobot~\citep{bu2025agibot}: moving downward while closing, moving upward while closing, moving downward while opening, and sweeping. We aggregate the transition features into clip-level representations, reduce them to 20 dimensions using PCA, and further project them into 3D using UMAP~\citep{mcinnes2018umap}. As shown in Fig.~\ref{fig:physical_language_structure}, the learned representations exhibit clear kinematic organization. In autonomous driving, stationary clips form a distinct cluster, while different steering patterns are arranged along a continuous manifold. In robotic manipulation, the four gripper transition patterns form compact and largely separated clusters. These results indicate that physical language captures meaningful state-transition semantics and organizes videos by how the world changes rather than by their visual content.



\begin{figure}[t]
    \centering
    \includegraphics[width=1.0\linewidth]{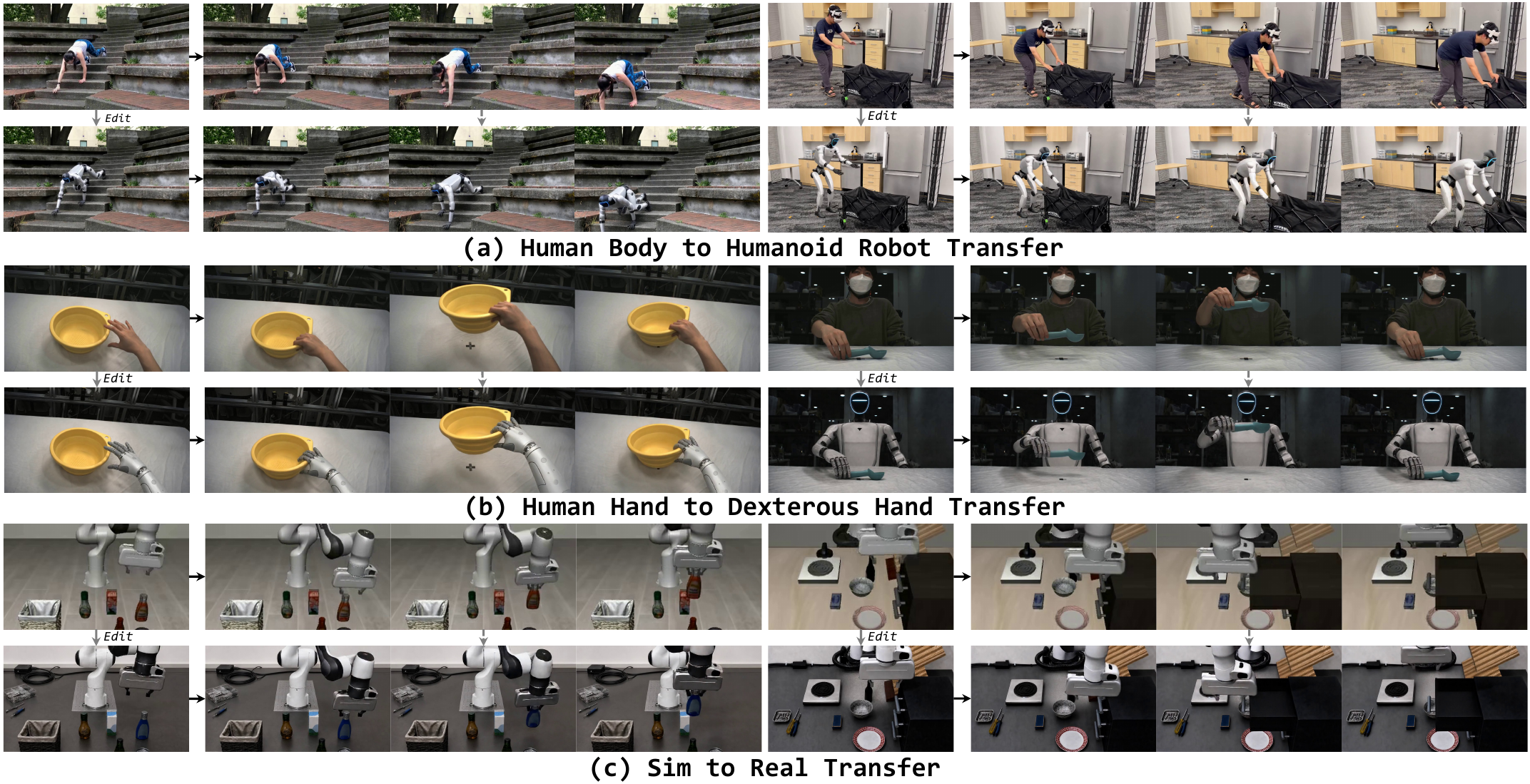}
    \vspace{-1.5em}
    \caption{\textbf{Zero-shot cross-embodiment and sim-to-real transfer with physical language.} We encode source state transitions into physical-language sequences and render them conditioned on edited first frames specifying new embodiments or visual domains, enabling zero-shot cross-embodiment and sim-to-real transfer.}
    \label{fig:physical_language_application}
\end{figure}

\subsection{Broader Applications}

We further demonstrate the broader application potential of \Methodname{}. By explicitly representing how the physical world evolves, physical language provides an appearance-disentangled interface for representing, controlling, and transferring state transitions across visual appearances and domains. This interface supports physically realistic video generation, fine-grained action-conditioned world modeling, interactive rollouts, and zero-shot transfer across embodiments and visual domains.

\paragraph{Physically Realistic Video World Model}

\Methodname{} can serve as a general video world model that predicts physically coherent future evolution from the current visual state. As shown in Fig.~\ref{fig:physical_generation}(a), it models diverse real-world dynamics, including ocean waves crashing against rocks, liquid pouring into a container, objects falling into hot oil and producing splashes, and a metal can exploding into flames and fragments. These examples capture both the complete temporal evolution of complex events and their fine-grained consequences, demonstrating the ability of \Methodname{} to model open-domain world dynamics with strong physical coherence.

\paragraph{Controllable and Interactive World Model}

Existing action-conditioned world models typically map control signals directly to future videos, making precise control over state evolution difficult. In contrast, \Methodname{} separates state-transition reasoning from pixel synthesis: given a trajectory or action signal, the Physical Language Reasoner first predicts the corresponding physical language, which is then decoded into a future video. We validate this capability on nuScenes~\citep{caesar2020nuscenes} and AGI-Bot RealRobot~\citep{bu2025agibot}. As shown in Fig.~\ref{fig:physical_generation}(b) and (c), \Methodname{} captures fine-grained variations in driving trajectories, such as different steering magnitudes, and accurately follows action signals to generate precise gripper movements. It also supports interactive world modeling under sequential control inputs. As shown in Fig.~\ref{fig:physical_generation}(d), the model updates the camera viewpoint and position according to successive controls while maintaining temporal consistency. These results demonstrate the potential of \Methodname{} for fine-grained, controllable, and interactive world modeling. Additional training and inference details for the controllable and interactive world models are provided in Appendix~\ref{app:controllable_interactive_world_model}.

\paragraph{Zero-Shot Cross-embodiment and Sim-to-real Transfer}

Because physical language disentangles state transitions from visual appearance, \Methodname{} supports zero-shot transfer across embodiments and visual domains. As shown in Fig.~\ref{fig:physical_language_application}(a) and (b), we encode the state transition of a human-motion video into physical language, edit its first frame by replacing the human body with a Unitree G1 humanoid or the human hand with a Sharpa dexterous hand, and decode the unchanged physical-language sequence conditioned on the edited first frame. This transfers full-body and hand motion patterns to substantially different embodiments without target-specific training. The same formulation also enables sim-to-real transfer. As shown in Fig.~\ref{fig:physical_language_application}(c), we transform the first frames of LIBERO~\citep{liu2023libero} videos into a realistic visual domain and render the original simulated state transitions under the new appearance. These results demonstrate that physical language provides a transferable interface across embodiments and visual domains, with potential for demonstration retargeting, cross-morphology transfer, and low-cost generation of realistic interaction data from simulation. Additional adaptation and inference details for zero-shot cross-embodiment and sim-to-real transfer are provided in Appendix~\ref{app:motion_transfer}.

\section{Conclusions}

We introduced \methodname{}, a physical world model built around physical language. Unlike conventional video models that predict future videos directly in pixel space, \methodname{} learns physical language: a compact discrete representation of state-transition patterns from unlabeled in-the-wild videos through self-supervised learning. Building on this representation, \methodname{} follows a reason-then-render paradigm: it first predicts future world evolution in the physical-language space and then renders the inferred transitions into videos. Extensive experiments across generation and understanding benchmarks demonstrate that \methodname{} generates physically coherent outcomes and identifies physically implausible events. Physical language further provides an interface for fine-grained action-conditioned simulation, interactive rollouts, and zero-shot motion transfer across embodiments and visual domains. These results highlight the potential of physical language for scalable, controllable, and transferable physical-world modeling. We further discuss limitations and future works in the Appendix~\ref{app:limitations}.

\clearpage
\appendix

\section{Additional Training Details}
\label{app:training_details}

\subsection{Physical Language Tokenizer}
\label{app:physical_language_tokenizer}

We train the Physical Language Tokenizer in two phases: large-scale pretraining followed by targeted SFT, using a joint temporal and spatial curriculum. For pretraining, we aggregate the datasets listed in Table~\ref{tab:pretrain_data}. After deduplication, we filter out clips with technical defects, including compression artifacts, corrupted frames, and watermarks, as well as those that fail the resolution or duration requirements or contain abrupt shot transitions. This process yields approximately 10K hours of unlabeled real-world videos for tokenizer pretraining.

\begin{table}[htbp]
    \caption{\textbf{Physical Language Tokenizer pretraining data.} We report the duration of each source corpus after data filtering. The mixture contains approximately 10K hours of unlabeled video.}
    \label{tab:pretrain_data}
    \centering
    \begin{minipage}{0.50\columnwidth}
    \centering
    \small
    \renewcommand{\arraystretch}{1.08}
    \begin{tabular*}{\linewidth}{@{\hspace{1em}\extracolsep{\fill}}lc@{\hspace{1em}}}
        \toprule
        \textbf{Dataset} & \textbf{Hours} \\
        \midrule
        In-house Data & 3329 \\
        HOIGen-1M~\citep{liu2025hoigen} & 2200 \\
        OpenVid-1M~\citep{nan2025openvid} & 2051 \\
        Moments in Time~\citep{monfort2019momentsintime} & 844 \\
        Kinetics-710~\citep{li2022Kinetics} & 840 \\
        Sekai-walking~\citep{li2026sekai} & 288 \\
        SSV2~\citep{goyal2017SSV2} & 245 \\
        UltraVideo~\citep{xue2025ultravideo} & 142 \\
        WISA-80K~\citep{wang2026wisa} & 141 \\
        \midrule
        \textbf{Overall} & \textbf{10K} \\
        \bottomrule
    \end{tabular*}
    \end{minipage}
\end{table}

During the first three pretraining stages, we keep both the input and reconstruction resolutions at $256 \times 448$ and progressively increase the clip duration from 1 second (9 frames) to 2 seconds (17 frames) and finally to 4 seconds (33 frames). In the fourth pretraining stage, the tokenizer continues to receive inputs at $256 \times 448$, while the diffusion decoder reconstructs targets at $512 \times 896$. This cross-resolution reconstruction encourages the compact physical-language bottleneck to focus on state transitions while allowing the diffusion decoder to rely on its generative prior to recover high-frequency appearance details.

For tokenizer SFT, we apply a stricter second-pass filtering procedure to the real-world corpus based on aesthetic quality, motion magnitude, and VLM-assessed state-transition observability. We further incorporate simulation videos after filtering out samples with low rendering quality, invalid duration, or insufficient object motion. The resulting corpus contains approximately 5M four-second clips, as summarized in Table~\ref{tab:sft_data}. We first fine-tune the full model on this curated corpus. In the final refinement stage, we freeze the spatiotemporal encoder, transition-level Q-Former, and FSQ quantizer, and optimize only the diffusion decoder. We additionally apply an entropy regularization loss~\citep{yu2023language} to encourage balanced utilization of the FSQ vocabulary and a REPA loss~\citep{yu2024representation} to improve the semantic quality of the diffusion representations. All training stages use 128 NVIDIA A100 GPUs. Table~\ref{tab:tokenizer_hyperparameters} summarizes the complete training schedule.

\begin{table}[htbp]
    \caption{\textbf{Physical Language Tokenizer fine-tuning data.} We distinguish between real-world and simulation sources and report the number of four-second clips contributed by each dataset.}
    \label{tab:sft_data}
    \centering
    \begin{minipage}{0.70\columnwidth}
    \centering
    \small
    \renewcommand{\arraystretch}{1.06}
    \begin{tabular*}{\linewidth}{@{\hspace{1em}\extracolsep{\fill}}lcc@{\hspace{1em}}}
        \toprule
        \textbf{Dataset} & \textbf{Type} & \textbf{Clips} \\
        \midrule
        In-house Data & Real & 2545K \\
        HOIGen-1M~\citep{liu2025hoigen} & Real & 574K \\
        Moments in Time~\citep{monfort2019momentsintime} & Real & 528K \\
        OpenVid-1M~\citep{nan2025openvid} & Real & 516K \\
        SSV2~\citep{goyal2017SSV2} & Real & 180K \\
        Sekai-walking~\citep{li2026sekai} & Real & 173K \\
        Kinetics-710~\citep{li2022Kinetics} & Real & 128K \\
        UltraVideo~\citep{xue2025ultravideo} & Real & 82K \\
        WISA-80K~\citep{wang2026wisa} & Real & 44K \\
        Physics101~\citep{wu2016physics101} & Simulation & 1K \\
        Phyco~\citep{narayanan2026phyco} & Simulation & 126K \\
        ComPhy~\citep{chen2022comphy} & Simulation & 60K \\
        Cosmos3~\citep{agarwal2026cosmos3} & Simulation & 51K \\
        CLEVRER~\citep{yi2019clevrer} & Simulation & 20K \\
        Physion++~\citep{tung2023physion++} & Simulation & 10K \\
        Physion~\citep{bear2021physion} & Simulation & 9K \\
        \midrule
        \textbf{Overall} & -- & \textbf{5M} \\
        \bottomrule
    \end{tabular*}
    \end{minipage}
\end{table}

\begin{table*}[htbp]
    \caption{\textbf{Physical Language Tokenizer training schedule.} Video and optimization settings for curriculum pretraining, full-model fine-tuning, and decoder-only refinement.}
    \label{tab:tokenizer_hyperparameters}
    \centering
    \begin{minipage}{0.96\textwidth}
    \centering
    \small
    \renewcommand{\arraystretch}{1.12}
    \resizebox{\linewidth}{!}{%
    \begin{tabular}{@{\hspace{0.5em}}l*{6}{c}@{\hspace{0.5em}}}
        \toprule
        \raisebox{0.5\baselineskip}[0pt][0pt]{\textbf{Configuration}} & \textbf{\shortstack{Pretrain\\Stage 1}} & \textbf{\shortstack{Pretrain\\Stage 2}} & \textbf{\shortstack{Pretrain\\Stage 3}} & \textbf{\shortstack{Pretrain\\Stage 4}} & \textbf{\shortstack{Full\\SFT}} & \textbf{\shortstack{Decoder-only\\SFT}} \\
        \midrule
        Frames & 9 & 17 & 33 & 33 & 33 & 33 \\
        Duration & 1s & 2s & 4s & 4s & 4s & 4s \\
        Global batch size & \multicolumn{6}{c}{256} \\
        Input resolution & \multicolumn{6}{c}{$256 \times 448$} \\
        Output resolution & $256 \times 448$ & $256 \times 448$ & $256 \times 448$ & $512 \times 896$ & $512 \times 896$ & $512 \times 896$ \\
        \midrule
        Training steps & 150K & 100K & 100K & 50K & 50K & 50K \\
        Initial pure-noise steps & 50K & 20K & 20K & 0 & 0 & 0 \\
        Warm-up steps & \multicolumn{6}{c}{1K} \\
        LoRA rank & \multicolumn{6}{c}{32} \\
        FSQ entropy-loss weight & \multicolumn{6}{c}{0.005} \\
        REPA-loss weight & \multicolumn{6}{c}{0.1} \\
        Learning rate & \multicolumn{6}{c}{$2 \times 10^{-5}$} \\
        AdamW betas & \multicolumn{6}{c}{$\beta_1=0.9,\;\beta_2=0.95$} \\
        Maximum gradient norm & \multicolumn{6}{c}{1.0} \\
        \bottomrule
    \end{tabular}%
    }
    \end{minipage}
\end{table*}

\subsection{Physical Language Reasoner}
\label{app:physical_language_reasoner}

We train the Physical Language Reasoner in two stages. In Stage 1, we perform continued pretraining on the same 5M four-second clips used for Physical Language Tokenizer SFT. For each clip, the first frame and a VLM-generated high-level action caption are provided as inputs, while the frozen Physical Language Tokenizer encodes the full video into the target physical-language sequence. In Stage 2, we perform motion-focused SFT on a corpus of approximately 1M clips, comprising 800K samples selected from the 5M-clip corpus using VLM-based motion-quality and physical-interaction filters and 200K additional simulator-generated samples. This stage concentrates training on salient and physically informative state transitions. The first frame is provided at a resolution of $512 \times 896$, and the action captions are generated using the prompt in Table~\ref{tab:caption_prompt}. Both training stages use 128 NVIDIA A100 GPUs, with their hyperparameters summarized in Table~\ref{tab:reasoner_hyperparameters}.

\begin{table}[htbp]
    \setlength{\belowcaptionskip}{5pt}
    \caption{\textbf{Physical Language Reasoner training settings.} Continued pretraining adapts the VLM to physical-language prediction, while SFT specializes it for physically informative state transitions.}
    \label{tab:reasoner_hyperparameters}
    \centering
    \begin{minipage}{0.69\columnwidth}
    \centering
    \small
    \renewcommand{\arraystretch}{1.10}
    \begin{tabularx}{\linewidth}{@{\hspace{1.2em}}l@{\hspace{0.2em}}*{2}{>{\centering\arraybackslash}X}@{\hspace{0.0em}}}
        \toprule
        \raisebox{4pt}[0pt][0pt]{\textbf{Configuration}} & \textbf{\shortstack{Continued\\Pretraining}} & \textbf{\shortstack{Motion-focused\\SFT}} \\
        \midrule
        Global batch size & \multicolumn{2}{c}{512} \\
        First-frame resolution & \multicolumn{2}{c}{$512 \times 896$} \\
        Training steps & 50K & 10K \\
        Warm-up steps & \multicolumn{2}{c}{1K} \\
        Learning rate & {$5 \times 10^{-5}$} & {$1 \times 10^{-5}$} \\
        AdamW betas & \multicolumn{2}{c}{$\beta_1=0.9,\;\beta_2=0.95$} \\
        Maximum gradient norm & \multicolumn{2}{c}{1.0} \\
        \bottomrule
    \end{tabularx}
    \end{minipage}
\end{table}

\begin{table*}[htbp]
    \caption{\textbf{Captioning prompt.} Prompt used to generate textual conditions for the four-second clips used to train the Physical Language Reasoner.}
    \label{tab:caption_prompt}
    \centering
    \begingroup
    \setlength{\fboxsep}{8pt}
    \fbox{%
        \begin{minipage}{0.95\textwidth}
            \small
            \raggedright
            \textbf{System prompt}

            \medskip
            {\ttfamily You are a professional video captioning expert. Your task is to generate a concise high-level action condition for the video clip, used to train a physical world model.

            \medskip
            Caption requirements:
            \begin{itemize}
                \setlength{\itemsep}{0pt}
                \setlength{\parskip}{0pt}
                \item Write one concise flowing English sentence or paragraph.
                \item Keep it 25-60 words.
                \item Use present tense and describe only visible content.
                \item Cover the main subject, action, setting, camera/framing, lighting, colors, and visual style when observable.
            \end{itemize}

            Critical rules:
            \begin{itemize}
                \setlength{\itemsep}{0pt}
                \setlength{\parskip}{0pt}
                \item Prioritize spatial precision.
                \item Focus on the high-level initiating action or interaction intent rather than narrating the full temporal evolution. Avoid fine-grained intermediate motions, state changes, and secondary physical effects.
            \end{itemize}}

            \medskip
            \textbf{User prompt}

            \medskip
            {\ttfamily Watch this video carefully and generate a caption strictly following your instructions.}
        \end{minipage}%
    }
    \endgroup
\end{table*}

\section{Additional Evaluation Details}

\subsection{Physical Video Generation Evaluation}
\label{app:physical_generation_evaluation}

We evaluate \Methodname{} on three physical video-generation benchmarks. For all benchmarks, the model takes the provided first frame and textual condition as input, and we follow the official evaluation protocols and released evaluators.

\paragraph{Physics-IQ Verified}
Physics-IQ Verified~\citep{radsch2026physicsiqverified} evaluates whether generated videos reproduce the outcomes of 66 controlled real-world physical experiments. The generated video is compared with the real reference using Spatial IoU (S-IoU), Spatiotemporal IoU (ST-IoU), and Weighted Spatial IoU (WS-IoU), which respectively evaluate the location, timing, and temporal frequency of scene changes. IQ-Score aggregates these metrics after normalization by the variation between repeated real experiments.

\paragraph{PhyGround}
PhyGround~\citep{lin2026phyground} contains 250 curated image--text conditions covering 13 physical laws in solid mechanics, fluid dynamics, and optics. We follow the official protocol using the released PhyJudge-9B evaluator and sample generated videos at 4 FPS. Each video is scored from 1 to 5 according to criterion-specific rubrics. General Quality evaluates prompt alignment, temporal validity, and object consistency, while Physics Score measures adherence to the physical laws applicable to each sample. Overall is the average of General Quality and Physics Score.

\paragraph{WorldModelBench}
WorldModelBench~\citep{li2026worldmodelbench} evaluates general world-modeling capability using 350 image--text conditions from seven domains and 56 subdomains, including autonomous driving, robotics, human activities, industrial scenes, gaming, animation, and natural environments. We follow the official protocol using the released 2B-VLM judge. Physics Adherence evaluates five physical-consistency criteria and ranges from 0 to 5, while Common Sense evaluates spatial and temporal visual quality and ranges from 0 to 2. The Total score is obtained by summing all metrics evaluated by the benchmark.

\subsection{Physical Video Understanding Evaluation}
\label{app:physical_understanding_evaluation}

\paragraph{IntPhys2}
IntPhys2~\citep{bordes2025intphys2} evaluates intuitive physics through matched possible and impossible videos involving object permanence, immutability, spatiotemporal continuity, and solidity. Its main set contains 1,012 videos organized into 253 quadruplets, with two possible and two impossible outcomes per scene. We compare the physical-language likelihoods of each matched possible--impossible pair and report the percentage of pairs for which the possible video receives the higher likelihood. Results are reported separately on the Easy, Medium, and Hard subsets, together with the Overall accuracy on the complete main set.

\paragraph{LikePhys}
LikePhys~\citep{yuan2025likephys} contains controlled valid--invalid video pairs from 12 scenarios. The videos within each comparison are matched in visual appearance and differ primarily in whether the relevant physical principles are satisfied. Following its likelihood-preference protocol, Plausibility Preference Error (PPE) is the percentage of comparisons for which an invalid video receives an equal or higher likelihood than a valid video; therefore, lower values indicate stronger physical understanding. We report PPE for rigid-body mechanics, fluid mechanics, and optical effects, as well as Avg.\ Error averaged over all 12 scenarios.

\paragraph{YoCausal}
YoCausal~\citep{xie2026yocausal} contains 1,232 real-world video pairs. Each natural forward video is paired with its temporally reversed counterpart, and the same caption is used for both directions. The Reversal Surprise Index (RSI) is the mean forward-video preference rate across the four source datasets, where a forward preference occurs when the natural video receives the higher physical-language likelihood; 50\% corresponds to chance. The benchmark further partitions videos into causal and non-causal subsets using VLM annotations and defines the Causality Cognition Index as $\mathrm{CCI}=\mathrm{RSI}(\mathcal{D}_{c})-\mathrm{RSI}(\mathcal{D}_{nc})$. Aggregate Rank combines the model rankings on RSI and CCI, with a lower rank indicating better overall performance.

\paragraph{Evaluation Protocol}
All three benchmarks are evaluated through pairwise likelihood comparison under a shared text condition. Following the official evaluation protocol, for LikePhys, we use the official prompt template provided for each scenario, which is shared by all valid and invalid variants of that scenario; for YoCausal, we use the benchmark-provided prompt for the forward video and its temporally reversed counterpart. IntPhys2 does not provide instance-level text descriptions. We therefore use a VLM to generate one scene-level caption for each video quadruplet. The VLM observes only the first frames of the four videos, without access to future frames or validity labels, and is instructed to describe only their shared visible scene and initial configuration. We also apply the resulting caption unchanged to all four videos in the quadruplet. Formally, let $(\mathbf{V}^{+},\mathbf{V}^{-})$ denote a matched pair and $c_{\mathrm{pair}}$ its shared text condition. For each video, we use its own first frame $I^{0}_{\pm}$ and the identical text condition $c_{\mathrm{pair}}$, encode it into a physical-language sequence $\mathbf{z}^{\pm}=\mathcal{T}_{\phi}(\mathbf{V}^{\pm})$, and compute
\begin{equation}
    s(\mathbf{V}^{\pm},c_{\mathrm{pair}})
    =
    \sum_{j=1}^{N}
    \log p_{\theta}
    \left(
        z^{\pm}_{j}
        \mid
        I^{0}_{\pm},
        c_{\mathrm{pair}},
        z^{\pm}_{<j}
    \right).
\end{equation}
Thus, the two likelihoods within each comparison differ only in the evaluated video and its corresponding first frame, while the text condition remains identical. The video with the higher sequence likelihood is selected as the physically valid or temporally natural one.

\section{Training and Inference Details for Broader Applications}
\label{app:broader_applications}

\subsection{Controllable and Interactive World Model}
\label{app:controllable_interactive_world_model}

\paragraph{Robotics World Model}
For robotic manipulation, we adapt \Methodname{} on the AGI-Bot RealRobot dataset~\citep{bu2025agibot}. We first fine-tune the Physical Language Tokenizer on RealRobot videos so that the learned physical language captures domain-specific robotic-arm and gripper transitions. For each four-second clip, the corresponding trajectories of all robot-arm joints and the gripper pose are sampled at 8 FPS. At every time step, these values are arranged in a fixed order and serialized into numerical text, and is used as the action condition for Physical Language Reasoner fine-tuning. At inference time, a desired joint and gripper trajectory is converted into the same textual format and provided together with the current observation. The reasoner translates this action sequence into physical language, and the domain-adapted decoder renders the resulting robotic interaction as a future video.

\paragraph{Driving World Model}
We follow the same procedure to construct the trajectory-conditioned driving world model on nuScenes~\citep{caesar2020nuscenes}. We first fine-tune the Physical Language Tokenizer on nuScenes videos to adapt the representation and decoder to driving-scene evolution. The future four-second ego-vehicle trajectory is sampled at 8 FPS and serialized into a temporally ordered numerical text sequence. The first frame and textualized ego trajectory are then used to fine-tune the Physical Language Reasoner, while the physical-language sequence extracted from the corresponding future video serves as the prediction target. During inference, a candidate ego trajectory is converted into the same numerical text format. The reasoner predicts how the scene should evolve under this trajectory in physical-language space, after which the decoder renders the corresponding future driving video. This formulation allows the model to follow fine-grained trajectory variations, including different steering directions and magnitudes, without modifying the general reason-then-render architecture.

\paragraph{Interactive World Model}
For interactive world modeling, control inputs are expressed directly in natural language as instructions describing the desired change in camera viewpoint or camera position. Since the current model generates four-second clips, we extend generation beyond this fixed horizon using a sliding-window autoregressive rollout. Starting from the current frame, the model predicts the next four-second video segment, and the final frame of the generated segment is subsequently used as the first-frame condition for the next window. Repeating this process produces longer videos and allows the camera trajectory to be modified progressively through sequential natural-language controls.

\subsection{Motion Transfer}
\label{app:motion_transfer}

\paragraph{Transfer Protocol}
Our motion-transfer pipeline does not require paired videos. Instead, we briefly fine-tune the Physical Language Tokenizer on videos from each source domain, enabling it to more accurately encode the corresponding motion patterns into physical language. After this source-domain adaptation, motion transfer can be performed directly without further training. Given a source video, we first encode its state transition into a physical-language sequence. We then use GPT-Image 2.0 to edit the first frame, replacing the original subject or visual appearance with the desired target embodiment or target domain. The unchanged physical-language sequence is subsequently decoded conditioned on the edited first frame. In this way, the first frame specifies what the target scene should look like, while physical language specifies how it should evolve over time.

\paragraph{Human-body Motion Transfer}
For human-body motion transfer, we construct a subset containing diverse and salient full-body human motions. The source videos are collected from ~\citep{chen2026warp,allshire2025visual}. We briefly fine-tune the Physical Language Tokenizer on this subset to adapt it to human-body motion. During inference, the source human-motion video is encoded into physical language, while its first frame is edited to replace the human subject with the target humanoid robot. Decoding the original physical-language sequence from the edited first frame transfers the full-body motion to the humanoid embodiment without paired human--robot training data.

\paragraph{Human-hand Motion Transfer}
For human-hand to dexterous-hand transfer, we construct a subset of human-hand videos with rich hand motions and object interactions from ~\citep{lim2026hrdexdb}. Although it provides paired human and robot data, we use only the human-hand videos for tokenizer adaptation and do not use the provided pairings or cross-embodiment correspondences. At inference time, we edit the first frame to replace the human hand with the target dexterous hand and decode the source physical-language sequence under this edited appearance condition. This transfers fine-grained hand motions and interaction patterns without paired supervision.

\paragraph{Sim-to-real Transfer}
For sim-to-real transfer, we briefly fine-tune the Physical Language Tokenizer on simulated interaction videos from LIBERO~\citep{liu2023libero}. No paired real-world videos are used during adaptation. Given a simulated source video, we encode its state transition into physical language and use GPT-Image 2.0 to transform the first frame into a realistic visual appearance. We then decode the unchanged physical-language sequence conditioned on the edited realistic frame, producing a realistic-looking video that preserves the interaction demonstrated in simulation.

\section{Additional Related Work}

\subsection{Latent-action World Models}
\label{sec:related_latent_action}

Latent-action models infer compact transition variables from observation sequences, typically through an inverse-dynamics bottleneck coupled with a forward predictor. Early approaches learn discrete latent actions from adjacent observations and use them to pretrain controllable world models or downstream policies~\citep{cui2023universal,schmidt2024LAPO,bruce2024genie,ye2025LAPA,gao2025adaworld}. Recent work extends this paradigm along several complementary directions. \citet{garrido2026learning} scale latent-action world models to heterogeneous in-the-wild videos and study constrained continuous representations. \citet{jiang2026olaf} align latent-action semantics across visual contexts through control-effect alignment, while \citet{wang2026factored} factorize scene dynamics into factor-wise latent actions for multi-entity environments. \citet{zhang2026dila} explicitly disentangle dynamics-relevant structure from visual content. Other methods connect latent actions to mixed-data control or unified modeling: \citet{alles2025latent} align action-free and action-conditioned trajectories for offline reinforcement learning, \citet{wang2026coevolving} jointly evolve latent-action inference and a pretrained video world model, and \citet{bi2025motus} integrate understanding, video generation, and action prediction within a unified architecture. \citet{chen2026unit} introduce a unified physical language that aligns human and humanoid actions through visually anchored cross-reconstruction for policy learning and world modeling. Despite these advances, latent-action world models are predominantly developed for control, often targeting specific embodiments and tasks or relatively constrained environments. In contrast, \Methodname{} uses this transition-modeling paradigm to represent the open-domain evolution of the physical world itself, learning a physical language from diverse in-the-wild videos and explicitly reasoning over it for both future-video generation and state-transition understanding.

\subsection{Video Tokenizers}

Video tokenizers compress videos into compact representations by exploiting spatial and temporal redundancy. A major line of work separates persistent visual content from temporal variation. \citet{sun2023moso} decomposes videos into scene, object, and motion tokens, while \citet{jin2024videolavit} interleaves keyframe and motion tokens. \citet{yu2024CMD,tian2025reducio,wang2025vidtwin,liu2025hi-vae} factorize videos through content frames, reference images, or hierarchical latent variables, and \citet{wang2026vtok,chen2026tivtok} further summarize shared appearance with keyframe or time-invariant tokens while compactly encoding temporal residuals. Recent work increasingly employs diffusion-based decoders, allowing compact representations to omit fine visual details that can be recovered by a generative prior. \citet{ge2025divot,li2024dicode} condition pretrained video diffusion models on compressed spatiotemporal features, while \citet{yang2025cdt} reconstructs videos with a conditional causal diffusion decoder. \citet{atanov2026videoflextok} learns variable-length coarse-to-fine tokens with a generative flow decoder, and \citet{teng2026adaptive} combines query-based 1D latents with a pixel-space diffusion decoder for adaptive compression. Closely related, \citet{ressler2025dismo} learns appearance-disentangled motion representations through video reconstruction for open-world motion transfer and motion understanding. Our Physical Language Tokenizer similarly adopts a generative decoder, but the learned physical language is not used solely for video reconstruction. It also serves as an explicit prediction target through which the model reasons about future world evolution before rendering the inferred transition into video.

\subsection{Physically Realistic Video Generation}

Existing approaches to improving the physical consistency of video generation broadly follow three directions. The first incorporates explicit constraints derived from physics simulators. For example, \citet{montanaro2024motioncraft} uses simulator-derived optical flow, \citet{liu2024physgen} performs rigid-body simulation, and \citet{foo2026PSIVG} integrates a 3D physics simulator. The second transfers physical priors from external models or simulated data, often through representation alignment~\citep{zhang2026videorepa,bhowmik2025moalign,kim2026TSA}. Along this line of work, \citet{xiong2026physalign} incorporates 3D geometric cues obtained from simulation, while \citet{jiang2026lamo} learns a motion prior from VAE latents. The third introduces explicit physical knowledge or intermediate reasoning into video generation. \citet{wang2026wisa,zhang2025DiffPhy} inject physical knowledge during training, whereas \citet{yang2025vlipp} uses a VLM to reason about object trajectories before generation. In contrast, \methodname{} does not rely on simulator-derived constraints, predefined physical variables, or explicit physical rules as supervisory signals. Instead, it learns physical language from large-scale unlabeled videos through self-supervision, abstracting patterns of state transition.

\section{Limitations and Future Work}
\label{app:limitations}

\paragraph{Limitations}

Despite the promising results, several limitations remain. First, physical language is currently learned as an empirical representation of state transitions rather than an explicit formulation of symbolic physical laws. This data-driven formulation enables scalable learning from diverse unlabeled videos, but the resulting discrete symbols are not yet directly grounded in interpretable physical variables or formal laws. Second, because physical language is learned primarily from observational videos, its coverage is constrained by what can be visually observed in the training data. World transitions that are visually ambiguous or largely inaccessible to vision, such as tactile interactions and microscopic particle dynamics, may therefore be more difficult to model. Finally, due to limitations in data and computational resources, our current implementation uses relatively small-scale models and a training corpus that remains limited relative to the diversity of the physical world. Nevertheless, the strong performance achieved at this scale suggests that larger models and more diverse training data could further improve the capabilities of physical-language-based world modeling.

\paragraph{Future Work}

Future work can extend physical language along several directions. First, beyond video generation, physical language may serve as an explicit intermediate representation for improving spatiotemporal understanding in VLMs and supporting more effective planning in embodied policies. Second, its transferability across appearances and embodiments opens a promising avenue for alleviating the scarcity of real-robot interaction data. In particular, state-transition patterns learned from large-scale human videos could potentially be transferred to robotic embodiments, enabling embodied systems to benefit from abundant human data with limited robot-specific supervision. Third, we plan to extend physical-language modeling beyond the current fixed-duration clips through hierarchical or recurrent prediction, enabling long-horizon world modeling while maintaining temporal consistency and faithfully tracking the evolving scene state. Finally, we will further scale \Methodname{} with stronger backbones, increased computational resources, and larger and more diverse training corpora to improve the performance of physical-language-based world modeling.




\clearpage

\bibliography{references}
\bibliographystyle{iclr2026_conference}

\end{document}